\documentclass{article} 
\usepackage[final]{colm2025_conference}

\usepackage{natbib}
\setcitestyle{square,comma,numbers,sort&compress}

\usepackage{url}
\usepackage{booktabs}
\usepackage{hyperref}
\usepackage{graphicx}
\usepackage{microtype}
\usepackage{multicol}
\usepackage{multirow}
\usepackage{lipsum}
\usepackage{enumitem}
\usepackage{pifont}
\usepackage{colortbl}
\usepackage{tabularx}
\usepackage{caption} 
\usepackage{float}
\usepackage{amsmath,amsfonts,amssymb,bbm}

\usepackage{bm}
\usepackage{lineno}
\usepackage{makecell}
\usepackage{array}
\usepackage{wrapfig}
\usepackage{tikz}
\usepackage[normalem]{ulem}
\usepackage{marvosym}
\usepackage{tcolorbox}
\tcbuselibrary{skins, breakable, theorems}
\usepackage{subcaption}  
\usepackage{float}  
\usepackage{listings}
\usepackage{fontawesome5}

\usepackage{tikz}
\usepackage{amsmath}
\usepackage{graphicx}
\usepackage{caption}
\usepackage{adjustbox}
\usetikzlibrary{3d, arrows.meta, positioning, calc}

\usepackage{lineno}

\usepackage[linesnumbered,ruled,vlined]{algorithm2e} 
\usepackage{amsmath}  
\usepackage{amssymb}  

\usepackage{CJKutf8}

\definecolor{1c}{RGB}{194, 213, 247}
\definecolor{2c}{RGB}{252, 225, 198}
\definecolor{3c}{RGB}{29, 108, 171}
\definecolor{4c}{RGB}{255, 116, 16}

\definecolor{darkred}{rgb}{0.7, 0, 0}
\definecolor{darkblue}{rgb}{0, 0., 0.7}
\definecolor{purple}{rgb}{0.7, 0, 0.7}
\definecolor{darkyellow}{rgb}{0.7, 0.5, 0}
\definecolor{darkgreen}{rgb}{0, 0.7, 0}

\newcommand{\order}{\textcolor{gray}{\texttt{order}}}

\newcommand{\character}{\textcolor{darkred}{\texttt{character}}}
\newcommand{\object}{\textcolor{purple}{\texttt{object}}}
\newcommand{\container}{\textcolor{darkblue}{\texttt{container}}}
\newcommand{\cc}{\textcolor{darkgreen}{\texttt{character-character}}}
\newcommand{\coc}{\textcolor{darkyellow}{\texttt{character-object-container}}}

\newcommand{\characters}{\textcolor{darkred}{\texttt{characters}}}
\newcommand{\objects}{\textcolor{purple}{\texttt{objects}}}
\newcommand{\containers}{\textcolor{darkblue}{\texttt{containers}}}

\hypersetup{colorlinks=true, citecolor=darkblue, linkcolor=darkblue, urlcolor=darkblue}

\definecolor{skyblue}{rgb}{0.75 0.88 0.98}
\definecolor{pink}{rgb}{1.0, 0.752, 0.796}

\usepackage{amsmath}

\title{\centering Spontaneous High-Order Generalization \\in Neural Theory-of-Mind Networks}

\author{
~~~~~~~~~~Yiming Wang \\
~~~~~~~~~~School of Computer Science\\
~~~~~~~~~~Shanghai Jiao Tong University\\
~~~~~~~~~~200240, Shanghai, P.R. China\\
~~~~~~~~~~\texttt{yiming.wang@sjtu.edu.cn}
\\ \And
~~~~~~~~~~Rui Wang \\
~~~~~~~~~~School of Computer Science\\
~~~~~~~~~~Shanghai Jiao Tong University\\
~~~~~~~~~~200240, Shanghai, P.R. China\\
~~~~~~~~~~\texttt{wangrui12@sjtu.edu.cn}
}

\begin{document}


\maketitle

\begin{abstract}
Theory-of-Mind (ToM) is a core human cognitive capacity for attributing mental states to self and others. Wimmer and Perner \citep{wimmer1983beliefs,perner1985john} demonstrated that humans progress from first- to higher-order ToM within a short span, completing this development before formal education or advanced skill acquisition. In contrast, neural networks represented by autoregressive language models progress from first- to higher-order ToM only alongside gains in advanced skills like reasoning, leaving open whether their trajectory can unfold independently, as in humans.
In this research, we provided evidence that neural networks could spontaneously generalize from first- to higher-order ToM without relying on advanced skills. We introduced a neural Theory-of-Mind network (ToMNN) that simulated a minimal cognitive system, acquiring only first-order ToM competence.
Evaluations of its second- and third-order ToM abilities showed accuracies well above chance.
Also, ToMNN exhibited a sharper decline when generalizing from first- to second-order ToM than from second- to higher orders, and its accuracy decreased with greater task complexity. These perceived difficulty patterns were aligned with human cognitive expectations.
Furthermore, the universality of results was confirmed across different parameter scales.
Our findings illuminate machine ToM generalization patterns and offer a foundation for developing more human-like cognitive systems.
\end{abstract}

\section{Introduction}

\begin{figure}[t]
  \centering
  \includegraphics[width=\linewidth]{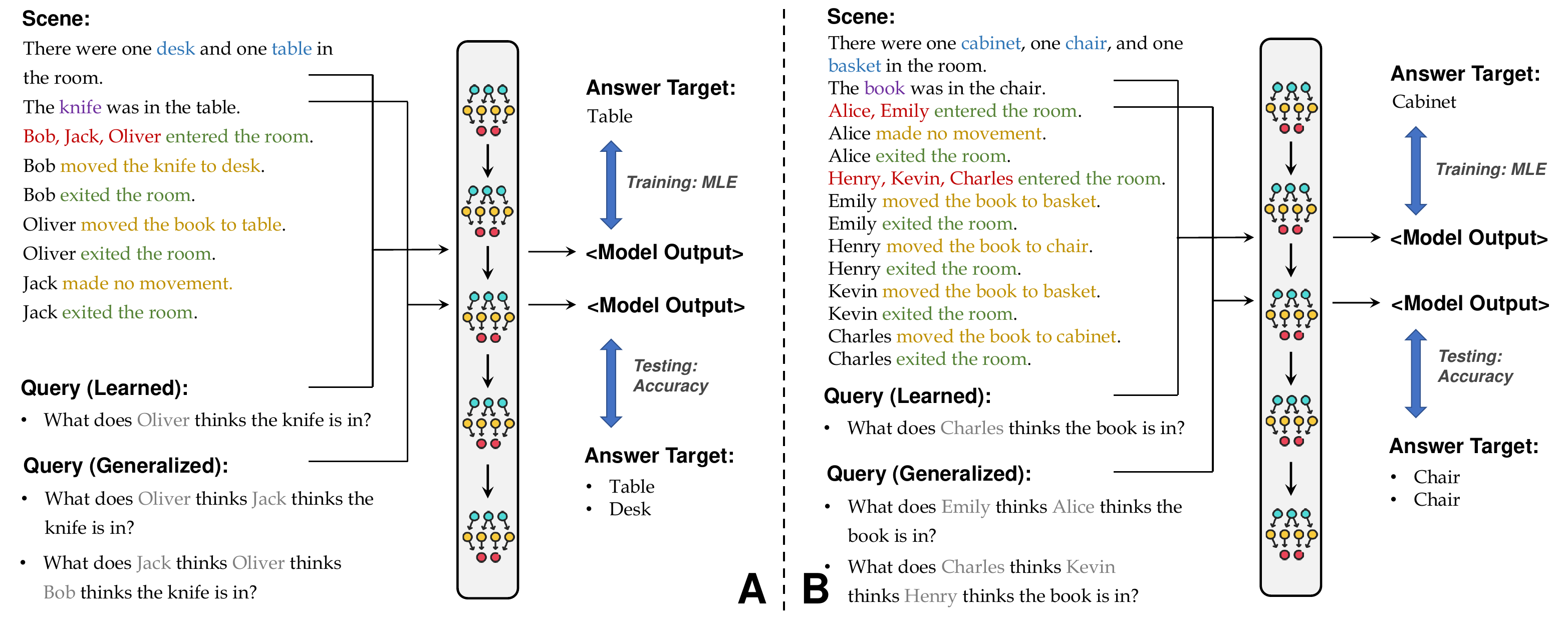}
  \caption{{\bf Implementation pipeline of ToMNN for learning and generalization via the Sally–Anne task.} {\bf (A)} A complete Sally-Anne task consists of a scene and a query. \textbf{During the \textit{learning} phase}, the model concurrently receives a background description (scene) and a first-order query (e.g., Oliver’s true belief). Its output is compared against the ground truth, and maximum likelihood estimation is performed for training. \textbf{During the \textit{generalization} phase}, the scene remains unchanged, but the query order is increased: the model must answer a {\it second-order} query (e.g., Oliver’s belief about Jack’s belief) and {\it third-order} query (e.g., Jack’s belief about Oliver’s belief about Bob’s belief). {\bf (B)} follows the same procedure as Episode {\bf (A)} but with increased scene complexity. In our experiments, task complexity is systematically controlled by categorizing scenes into distinct complexity levels and constructing parallel experimental groups to ensure robust generalization evaluation.}
  \label{fig:first}
\end{figure}

Humans are adept at perceiving their own and others’ emotions and cognition, as well as inferring mental states in social contexts. This set of cognitive skills is collectively referred to as the Theory-of-Mind (ToM) \cite{premack1978does,leslie1987pretense,tager2000componential,beaudoin2020systematic}, which emerges early in development, with infants as young as 14 months already showing sensitivity to others’ perceptions \citep{gopnik1997words,onishi200515}.
Over four decades ago, Wimmer and Perner identified key milestones in ToM development. By ages 4 to 6, children can reliably represent and reason about an individual’s belief state, like ``sb. thinks...'', marking the onset of first-order ToM \citep{wimmer1983beliefs}. Further, by ages 6 to 7, this extends to socially embedded contexts, enabling children to capture nested beliefs, like ``sb. thinks that sb. thinks (maybe more sb.)... '', signaling the emergence of second- and higher-order ToM \citep{perner1985john}.
The progression from first-order to higher-order ToM in humans exhibits several distinctive features. It unfolds sequentially with increasing orders rather than emerging simultaneously. More importantly, it occurs in remarkably short time and is largely completed before formal education \citep{sullivan1994preschoolers} or advanced skill acquisition, such as explicit reasoning \citep{wellman1990child,astington1993child,leslie1994tomm,wellman2004scaling,carpendale2004constructing}.

On the computer science side, scientists have explored neural networks with sophisticated architectures to effectively model human behavior \citep{bengio2003neural,mikolov2010recurrent}. Among them, autoregressive language models have emerged as the most promising paradigm \citep{radford2018improving}, flourishing in recent years and pursuing artificial general intelligence (AGI) across domains such as language, reasoning, and decision-making \citep{radford2019language,brown2020language,achiam2023gpt,guo2025deepseek,xu2025information}.
However, unlike humans, even powerful language models show no systematic evidence of possessing only first-order ToM in their early evolution stages. Signs of first-order ToM emerge only once models reach massive parameter scales and undergo extensive pretraining --- by which point they have already acquired substantial world knowledge and complex reasoning capacities \citep{kosinski2023theory,wang2024meta}. As these models continue to evolve, their performance on higher-order ToM tasks also improves, yet this progress consistently coincides with parallel gains in knowledge and reasoning \citep{strachan2024testing}.

Viewed in this light, neural networks that exhibit first-order ToM appear to progress toward higher-order ToM only in tandem with the acquisition of advanced skills. This trajectory contrasts with the developmental pattern identified by Wimmer and Perner in humans \citep{wimmer1983beliefs,perner1985john}. Because networks already possess extensive world knowledge and complex reasoning when first-order ToM emerges, disentangling these factors to replicate the human trajectory is no longer feasible. Consequently, whether the progression from first- to higher-order ToM in neural networks fundamentally differs from that in humans --- requiring the intervention of advanced skills --- becomes an open debate.

\begin{figure}[t]
  \centering
  \includegraphics[width=1\columnwidth]{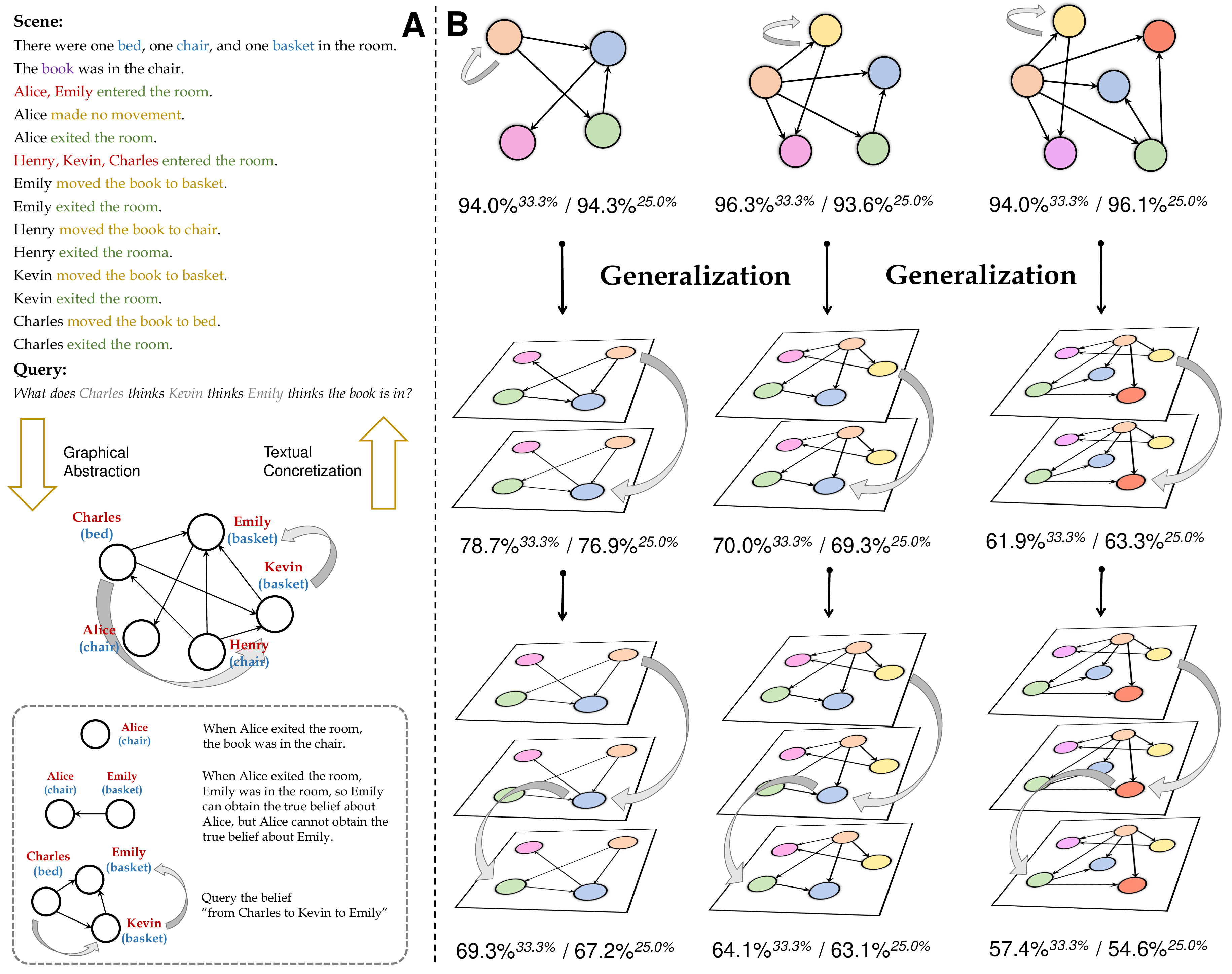}
  \caption{{\bf Sally-Anne task complexity control and experimental group construction across different task complexities.} {\bf (A)} The scene of the Sally-Anne task comprises multiple static and dynamic elements, which determine the task complexity. The textual scene can be losslessly abstracted as a graph structure, and the graph can be reversely reconstructed into its textual form. Graph-scale variables (e.g., number of nodes and edges) can be mapped to the scene variables, enabling precise complexity control. {\bf (B)} Generalization is evaluated on parallel experimental groups across different complexity levels. From left to right, three complexity settings are shown; within each group, rows indicate (top to bottom) first-order learning accuracy, and second- and third-order generalization accuracy. Superscripts denote the expected random accuracy for each setting.
}
  \label{fig:evaluation}
\end{figure}

In this research, we introduce a neural Theory-of-Mind network (ToMNN) that simulates a minimal cognitive system, acquiring only first-order ToM competence without incorporating any additional skills. We present preliminary evidence that \textit{\textbf{neural networks can spontaneously generalize from first- to higher-order ToM in the absence of any advanced skills}}, thereby mirroring the developmental patterns observed in humans.
Specifically, we implemented the simulation using modern transformer-based autoregressive language models \citep{vaswani2017attention} to ensure the most effective acquisition of first-order ToM.
To form a closed-loop research, we adopted the Sally-Anne task originally introduced by Wimmer and Perner in their human study \citep{wimmer1983beliefs,perner1985john} for ToMNN learning, which is a canonical paradigm for abstract ToM orders (see Figure \ref{fig:first}).
We strictly controlled the task complexity to ensure that generalization across ToM orders was evaluated under fully matched distributions of complexity. Based on this design, multiple experimental groups were constructed, enhancing the robustness of our conclusions (see Figure \ref{fig:evaluation}).

Through our ToMNN, we found that when neural networks acquire only first-order ToM without possessing any additional abilities, they exhibit the potential to generalize to higher-order ToM spontaneously. This phenomenon was consistently validated across different levels of task complexity (results of Section \ref{sec:results-main}).
Furthermore, ToMNN showed a marked drop in accuracy when generalizing from first- to second-order ToM, whereas the decline was much smaller when extending to third-order ToM (results of Section \ref{sec:results-preception}). For humans, the first-to-second-order transition marks a qualitative milestone --- from understanding others' mental states to recursively reasoning about others’ beliefs \citep{apperly2010mindreaders,miller2012theory}, while higher orders primarily add reasoning complexity without altering cognitive capacity, making the second-to-third-order transition far less demanding.
The generalization pattern in ToMNN thus not only demonstrates its potential to generalize but also indicates that its perception of ToM difficulty aligns with human developmental patterns.

\section{Results}

\begin{figure}[htbp]
  \centering
  \includegraphics[width=1\linewidth]{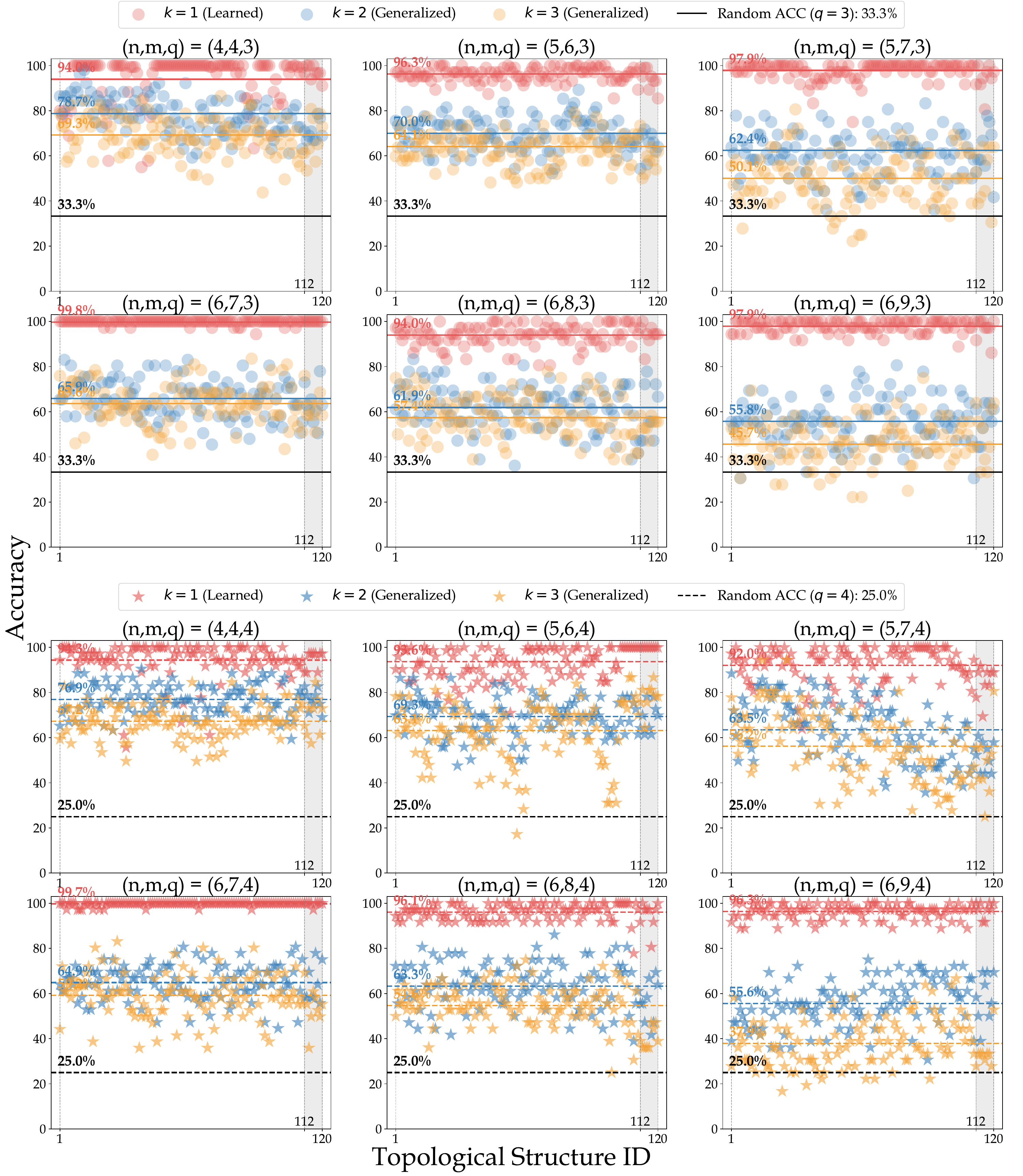}
  \caption{{\bf ToMNN accuracies in learning and generalization settings.} Results from 12 parallel evaluation groups with varying configurations determined by the three complexity variables $(n, m, q)$ are presented. Each group contains 120 distinct graph structures abstracted from task scenes, with 112 structures included in the training data and the remaining 8 held out for testing (shaded in gray). In each sub-figure, red points denote ToMNN’s learning accuracy on first-order ToM, while blue and yellow points indicate its generalization accuracy on second- and third-order ToM, respectively. Each graph structure corresponds to a scene template enriched with diverse semantic instantiations, generating a large number of tasks. Accordingly, the accuracy of each point represents the average performance across the tasks associated with that structure. The colored lines represent mean accuracies for each order, and the black line indicates the random baseline accuracy ($1/q$).}
  \label{fig:order-results}
\vspace{-0.05in}
\end{figure}

We implement ToMNN using the Sally–Anne task, as shown in Figure \ref{fig:first}. Each task consists of two components: the {\bf Scene} and the {\bf Query}, with the latter specifying the ToM order. ToMNN adopts a transformer-based decoder-only architecture \citep{vaswani2017attention} to acquire first-order ToM and then generalizes to higher-order ToM (see Section \ref{sec:model} for architectural details). Throughout this process, we assess whether ToMNN has fully learned first-order ToM and the extent to which it achieves generalization accuracy on higher-order ToM.

The scene in the Sally–Anne task consists of multiple \characters, \objects, and \containers ~engaged in \cc ~and \coc ~interactions in temporal order, with the scale of these variables directly determining task complexity. To eliminate potential confounds in our generalization evaluation, we quantified all complexity factors and organized parallel experimental groups, each controlling for complexity within the same distribution, as shown in Figure \ref{fig:evaluation}.
Each scene was losslessly mapped to a graph representation, where the number of nodes $n$, edges $m$, and node attribute types $q$ could be explicitly manipulated to regulate complexity.
The detailed task complexity control procedure based on graph abstraction is described in Section \ref{sec:abstraction}.
Then, we constructed independent experimental groups with fixed variable sizes $(n, m, q)$ to ensure that task complexity remained constant within each group, meaning that scene complexity was rigorously matched between learning and generalization.
The detailed procedure for constructing the experimental group is provided in Section \ref{sec:data} (dataset construction via the textual concretization) and Section \ref{sec:protocol} (evaluation protocol).

\subsection{ToMNN can Spontaneously Generalize Higher-Order ToM}
\label{sec:results-main}

Among the variables $(n,m,q)$, $q$ governs both the complexity of the scene and the query’s search space. Answering a query is essentially equivalent to selecting the \container ~that satisfies the condition from among $q$ candidates. Thus, in a setting with $q$ \containers, the random accuracy is $1/q$. This expected baseline serves as a critical threshold for assessing whether the model demonstrates genuine generalizability.

Figure \ref{fig:order-results} presents the results of 12 experimental groups under different $(n, m, q)$ configurations. Notably, ToMNN’s learning performance on first-order ToM ($k=1$) converged to nearly 100\% accuracy, confirming its successful acquisition of first-order ToM and providing a reliable basis for subsequent generalization analyses.
When tested in second-order ($k=2$) and third-order ($k=3$) ToM, performance decreased and exhibited greater variance between different graph structures. However, the results did not fluctuate around the random baseline ($1/q$), but instead remained consistently and substantially above it.
Specifically, when $q=3$ (random baseline = 33.3\%), the average generalization accuracy for second-order ToM was 2.36, 2.10, 1.87, 1.98, 1.86, and 1.68 times the random baseline in six $(n,m)$ settings, while third-order ToM achieved 2.08, 1.92, 1.50, 1.91, 1.72, and 1.37 times the random baseline.
When $q=4$ (random baseline = 25.0\%), the same relative pattern was observed: second-order ToM maintained multiples of 3.08, 2.77, 2.54, 2.60, 2.53, and 2.22, while third-order ToM showed multiples of 2.69, 2.52, 2.25, 2.37, 2.18, and 1.52.
This suggested that the ToMNN had already developed a non-trivial capacity for higher-order ToM generalization, although it was not achieving perfect generalization. Furthermore, this effect consistently emerges in the 12 experimental groups, ensuring the robustness of the generalization conclusions.

\subsection{ToMNN’s Perception of Generalization Difficulty Aligns with Human Cognition}
\label{sec:results-preception}

We observe that the most substantial decline in accuracy emerges between $k = 1$ and $k = 2$, whereas the subsequent decrease at $k = 3$ is comparatively moderate. Specifically, when $q = 3$, across six different $(n, m)$ configurations, the accuracy drop from $k = 1$ to $k = 2$ was 15.3\%, 26.3\%, 35.5\%, 33.9\%, 32.1\%, and 42.1\%. By contrast, the additional decline from $k = 2$ to $k = 3$ amounted to only 9.4\%, 5.9\%, 12.3\%, 2.3\%, 4.5\%, and 10.1\%.
This non-linear degradation suggests that the transition from first- to second-order reasoning constitutes a qualitative, rather than merely quantitative, enhancement in cognitive complexity.

From a cognitive science perspective, this pattern aligns with established findings on human ToM development: first-order ToM primarily involves attributing mental states to individuals, whereas higher-order ToM requires recursively modeling the beliefs of others --- progressing from simple {\bf state attribution} to {\bf state reasoning} \citep{perner1985john, apperly2010mindreaders, miller2012theory}.
Accordingly, the transition from first- to second-order ToM constitutes a qualitative leap --- from mere understanding of others’ beliefs to reasoning about them --- whereas higher orders primarily add complexity without altering the underlying cognitive capacity. The marked performance drop at $k = 2$ thus likely signifies the developmental threshold at which humans acquire second-order ToM \citep{perner1999development}.

\begin{figure}[t]
  \centering
  \includegraphics[width=1\linewidth]{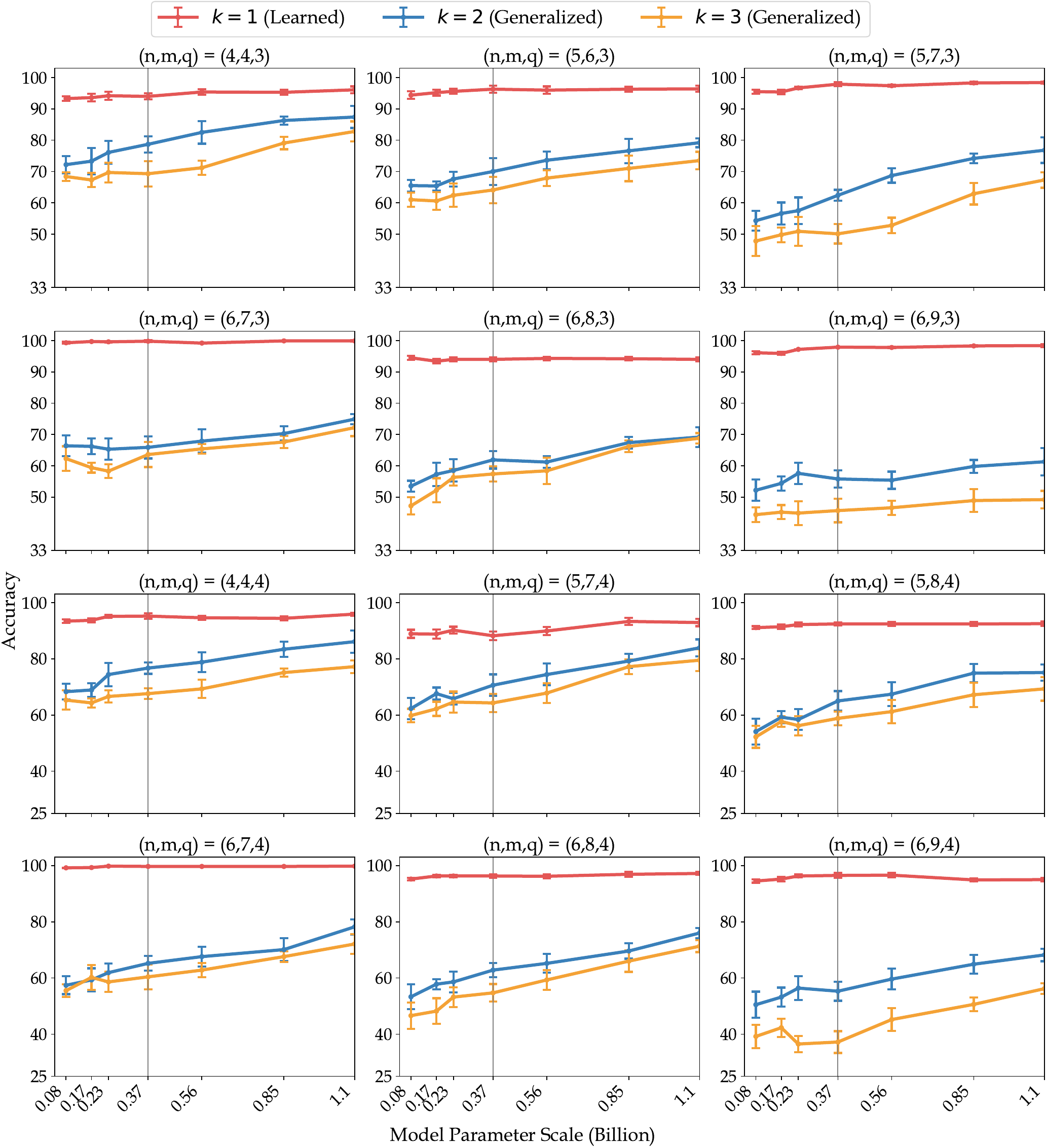}
  \caption{{\bf ToMNN accuracies in learning and generalization across model parameter sizes.} Each subplot shows a specific $(n,m,q)$ configuration, with the $x$-axis denoting parameter size and the $y$-axis accuracy. The red line indicates learning performance on first-order ToM ($k=1$), while the blue and yellow lines show generalization performances to second- and third-order ToM ($k=2,3$).}
  \label{fig:model-scale}
\vspace{-0.05in}
\end{figure}

\subsection{Task Complexity Influence the Generalization Difficulty}

We also examined how task complexity influenced generalization ability.
First, increasing $m$ while holding $n$ fixed, {\it i.e.}, introducing more \cc ~interactions, consistently reduced generalization performance. For example, when $q = 3$ and $n = 6$, second-order accuracy decreased from 65.9\% $\rightarrow$ 61.9\% $\rightarrow$ 55.8\% as $m$ increased from 7 $\rightarrow$ 8 $\rightarrow$ 9, while third-order accuracy declined from 63.6\% $\rightarrow$ 57.4\% $\rightarrow$ 45.7\%. 
Second, increasing $n$ while keeping graph density largely unchanged (for $(n,m) = (4,4), (5,6), (6,7)$, with all densities within 0.6–0.7), {\it i.e.}, adding more \characters, also lowered performance. For example, with $q = 3$, second-order accuracy dropped from 78.7\% to 70.0\% to 65.9\% as $n$ increased from 4 $\rightarrow$ 5 $\rightarrow$ 6, and third-order accuracy decreased from 69.3\% $\rightarrow$ 64.1\%  $\rightarrow$ 63.6\%.

We also examined whether option complexity, controlled by $q$, affects accuracy under the same $(n, m)$ settings. When $q = 3$, the random baseline accuracy is 1.33 times that of $q = 4$. However, for $k = 2$, the accuracies with $q = 3$ are only 1.02, 1.01, 0.98, 1.02, 0.98, and 1.00 times those with $q = 4$ in six different configurations $(n, m)$. Similarly, for $k = 3$, the ratios are 1.03, 1.02, 0.89, 1.07, 1.05, and 1.23. These results indicate that, although the search space expands with larger $q$, the model's generalization accuracy remains nearly unchanged. This suggests that generalization ability is largely insensitive to the option complexity $q$.

\subsection{The Generalization Ability is General Across Model Scales}

We also varied the model parameter scale to assess whether the observed generalization phenomenon holds consistently across different model sizes. This was done by adjusting the \texttt{hidden dimension}, \texttt{layer number}, and \texttt{attention head number}, as shown in Table \ref{tab:model-scale-parameter}, with the results presented in Figure \ref{fig:model-scale}.

\begin{wraptable}{r}{0.46\textwidth}
\vspace{-0.17in}
\caption{{\bf Model Scale Ablation Setup:} parameter sizes and implementation details.}
\vspace{-0.1in}
\centering
\footnotesize
\renewcommand\arraystretch{1.0}
  
\setlength{\tabcolsep}{1mm}{
\resizebox{0.46\textwidth}{!}{
\begin{tabular}{m{0.12\textwidth}m{0.14\textwidth}m{0.1\textwidth}m{0.1\textwidth}}

\toprule

\makecell{{\bf Parameter} \\ {\bf Size}} &
\makecell{{\bf Hidden} \\ {\bf Dimension}} &
\makecell{{\bf Layer} \\ {\bf Number}} &
\makecell{{\bf Head} \\ {\bf Number}}
\\
\midrule

\multicolumn{1}{c}{0.08B} & \multicolumn{1}{c}{512} & \multicolumn{1}{c}{12} & \multicolumn{1}{c}{8} \\
\multicolumn{1}{c}{0.17B} & \multicolumn{1}{c}{512} & \multicolumn{1}{c}{16} & \multicolumn{1}{c}{8} \\
\multicolumn{1}{c}{0.23B} & \multicolumn{1}{c}{1024} & \multicolumn{1}{c}{16} & \multicolumn{1}{c}{16} \\

\rowcolor{gray!20}
\multicolumn{1}{c}{0.37B} & \multicolumn{1}{c}{1024} & \multicolumn{1}{c}{24} & \multicolumn{1}{c}{16} \\

\multicolumn{1}{c}{0.56B} & \multicolumn{1}{c}{1280} & \multicolumn{1}{c}{24} & \multicolumn{1}{c}{20} \\
\multicolumn{1}{c}{0.85B} & \multicolumn{1}{c}{1600} & \multicolumn{1}{c}{24} & \multicolumn{1}{c}{25} \\
\multicolumn{1}{c}{1.10B} & \multicolumn{1}{c}{1600} & \multicolumn{1}{c}{32} & \multicolumn{1}{c}{25} \\

\bottomrule

\end{tabular}%
}
}

\label{tab:model-scale-parameter}%
\vspace{-0.2in}
\end{wraptable}

Using the 0.37B parameter model from the main experiment in Figure \ref{fig:order-results} as the baseline, we first scaled up the model in two steps, with the largest configuration surpassing 1B parameters. As the parameter size increased, generalization accuracy continued to improve, but gradually plateaued. On the ~1B scale, the accuracy for the second- and third-order queries stabilized above 80\%. This matches the expectation that larger model capacity, by storing more information, can capture more complex patterns conducive to ToM generalization.

Although larger models tended to generalize better, determining the lower bound of the model size required for such a capability is less predictable. To explore this, we scaled down below 0.37B in three steps, with the smallest model containing just 0.08B parameters, which is already smaller than the earliest pretrained language models such as BERT \citep{devlin2019bert} and GPT-1 \citep{radford2018improving}. Surprisingly, generalization accuracy remained relatively stable even at this smallest scale, with no significant drop observed across all 12 evaluation groups. This indicates that even tiny transformer-based language models can possess substantial potential for higher-order ToM generalization.

\section{Discussion}
\label{sec:discussion}

Over the past four decades, psychologists and cognitive scientists have extensively examined the developmental trajectory of human ToM, tracing its emergence from infancy to adulthood, with each stage characterized by distinct milestones and developmental pathways. In contrast, neural language models are a much more recent development, and their opacity has made it difficult to provide systematic interpretations of their ``mental states'' or to establish a credible account of their developmental course.
Nonetheless, the question of whether language models can capture human-like cognitive abilities has long been at the center of interdisciplinary debate \citep{lake2017building}. Critics contend that language models lack innate linguistic and cognitive priors and thus cannot genuinely capture the essence of human language and thought, while proponents argue that large-scale data-driven learning may spontaneously give rise to human-like reasoning patterns. To move this debate forward, and to better understand the similarities and differences in ToM development between humans and machines, it is crucial to place their performances in direct, parallel comparison.

By introducing ToMNN, we systematically addressed this challenge. Building on the human studies conducted by Wimmer and Perner four decades ago \citep{wimmer1983beliefs,perner1985john}, we reverse-engineered human inductive biases within a neural network architecture. We employed an autoregressive language model architecture to maximize the expressive power of neural networks, while synthesizing large-scale first-order ToM data to simulate priors, constraining the network to learn solely from them to minimize its cognitive capability space.
ToMNN demonstrated that neural networks, even without acquiring advanced skills and equipped solely with first-order ToM, still possessed substantial potential for spontaneous generalization to higher-order ToM. Moreover, its generalization accuracy across different orders and task complexities aligned with human cognitive expectations regarding the perceived difficulty of these scenarios. 
These findings in our ToMNN provided a clearer understanding of the similarities and differences between humans and machines in ToM order generalization patterns.

Although the practical success of ToMNN provided systematic evidence on the open question of neural networks’ potential for ToM order generalization, certain limitations remained. ToM beyond the third order often requires extensive specialized training even in adults and thus lies outside the scope of early human development \citep{kinderman1998theory,dunbar2003social}. For this reason, we excluded constructions above the third order. However, this truncation enhances the relevance of our comparisons and conclusions, as it confines the analysis to the range of ToM abilities that humans acquire without advanced skill training.
In addition, the current implementation of ToMNN restricts our generalization results to the linguistic modality, leaving open the question of whether similar evidence could be obtained in the visual domain. However, since transformer-based autoregressive language models are the most widely adopted paradigm, our findings provide meaningful contributions to this line of research. Future work could map textual scenes into image-dominant representations and apply the ToMNN framework within architectures equipped with visual encoders \citep{krizhevsky2012imagenet,radford2021learning}, thereby enabling cross-modal investigation.

Back to the broader debate on the similarities and differences in ToM development between humans and machines, this study contributes a preliminary piece to the large puzzle: under order-generalization, both exhibit spontaneous patterns of generalization. Yet, to advance this debate with more conclusive evidence, deeper developmental dynamics must be examined. For instance, human low-order ToM can emerge in relatively simple linguistic environments; identifying the minimal conditions under which models acquire first-order ToM --- by reducing inductive bias and defining the minimal corpus boundary—would help uncover the shared starting point of human and machine ToM, while offering comparative evidence for early ToM development in children. Moreover, whereas humans typically establish higher-order ToM before acquiring more complex skills such as logic, mathematics, and planning, it remains unclear whether machines follow a comparable learning trajectory. Do humans first construct the foundation of nested representations through low-order ToM before extending to more advanced compositional abilities? And do machines reveal analogous developmental pathways? Addressing these questions will extend the present findings, yielding stronger evidence for human–machine cognitive alignment and helping to map a continuum from low-level social cognition to higher-order abstract reasoning.

\section{Method}
\label{sec:deconstruction}

\subsection{Overview: Task Deconstruction}
\label{sec:deconstruction}










Our analyses and experiments build upon the classic Sally-Anne task, a well-established paradigm for assessing a critical developmental milestone in human ToM \citep{wimmer1983beliefs, baron1985does, perner1989exploration, wellman2001meta}. A complete Sally–Anne task comprises two components: a {\bf Scene} and a {\bf Query}, as illustrated in Figure \ref{fig:first}.
In this paradigm, the subject, observing the scene from an omniscient perspective, follows the sequence of events, infers how each character attributes beliefs to others, and ultimately responds to the query. For analytical clarity, we further decompose these two components into static and dynamic elements as follows:
\begin{itemize}[leftmargin=0.6cm]
    \item The scene comprises three types of static elements: \character, \object, and \container. As events unfold sequentially, two types of dynamic interactions occur: character entry-exit actions (\cc), and the placement of objects into containers by characters (\coc). The richer these elements are, the higher the scene complexity becomes.
    \item The query focuses on belief flow between characters, with the key factor being \order, the number of nested beliefs. For example, ``{\it What does A think...?}'' is first order, while ``{\it What does A think B thinks...?}'' is second order. Higher-order tasks require more reasoning steps, increasing the cognitive complexity for the subject.
\end{itemize}

We illustrate the specific forms of all static elements and dynamic interactions in the example shown in the left part of Figure \ref{fig:evaluation}.
Our modeling and experimental design are carried out within a {\bf fully controllable} Sally-Anne task setting. Due to the unstructured nature of textual data, we losslessly map the textual information in the Sally-Anne task into a graph structure, {\bf enabling structured control over all involved attributes}. This allows for noise-free generalization analysis.
In our experiments, for each of the $N$ attributes, we fix $(N - 1)$ of them and investigate the model’s generalization ability along the remaining attribute.

\subsection{Graphical Abstraction}
\label{sec:abstraction}

The textual form of the Sally-Anne task obscures the structured elements and dynamic interactions, complicating efforts to systematically control variables.
To address this, we propose a graphical abstraction method. Specifically, a complete Sally-Anne task scene is first losslessly mapped from text into a belief graph $\mathcal{G}$, which captures all three types of static elements and two types of dynamic interactions. The query is then mapped into a belief flow $\mathcal{F}$.
An example is illustrated on the left side of Figure \ref{fig:evaluation}.

\paragraph{Scene: Belief Graph}
\

Assume the scene involves $n$ \characters ~$\mathcal{C} = \{c_i\}_{i=1}^n$, $q$ \containers ~$\mathcal{B} = \{b_i\}_{i=1}^q$, and a single \object ~$\mathcal{A} = \{a\}$. All \characters, \containers, and the \object ~are represented by meaningful vocabularies sampled from a closed language space $\mathcal{L}$, so $a, b_i, c_i \in \mathcal{L}$.
In the scene, \characters ~are core, with \objects ~and \containers ~directly associated with it.
Dynamic interactions generate two critical information: 
(i) Only when a \character ~exits the scene are \cc ~interactions triggered, which influence other characters' beliefs about that character;
(ii) At the same time, changes in the relative position of the \object ~and \containers ~trigger the \coc ~interactions, which determine the character's true belief to the environment.
We construct a belief graph $\mathcal{G} = (\mathcal{V}, \mathcal{E})$ with \characters ~as the node set $\mathcal{V}$ and treat \objects ~and \containers ~as attributes of these nodes, following the below rules that can reflect the above critical information:

\begin{itemize}[leftmargin=0.6cm]
    \item {\bf Node:} The set of all \characters ~constitutes the original node set. For node attributes, when a character exits the scene, they may either place the \object ~$a$ into a random \container ~$b_{i'}$ or leave it unmoved, producing the \coc ~interaction. Each \character ~node $c_i$ thus records a \container ~$b_{i'}$, representing its belief about the location of $a$. Formally, $$\bm{\mathcal{V} = \{c_i \circ b_{i'} \mid b_{i'} \in \mathcal{B}\}_{i=1}^n},$$ where $n = |\mathcal{V}|$ denotes the node complexity, corresponding to the number of \characters.
    \item {\bf Edge:} When a \character ~$c_i$ exits the scene, a \cc ~interaction occurs: every \character ~still in the scene $\{c_{i_j}\}_j$ forms a directed edge toward $c_i$, {\it i.e.}, $(c_{i_j}, c_i)$. This indicates that $c_{i_j}$ can hold a true belief about $c_i$, but not vice versa. Formally, $$\bm{\mathcal{E} = \{\,(c_{i_j}, c_i)\,\}_j^{\,n}},$$ where $m = |\mathcal{E}|$ denotes the edge complexity, corresponding to the number of \cc ~interactions.
\end{itemize}

\paragraph{Query: Belief Flow}
\

After observing the scene, the subject is asked a query of the form:
\begin{center}
{\it What does $c'_1$ thinks $c'_2$ thinks $\cdots$ $c'_k$ think the [\object] is in (which [\container])?}
\end{center}
Here, $k$ denotes the reasoning \order, with all $c'_i \in \mathcal{C}$ distinct, so $k \leq n$. This query defines a belief flow $\mathcal{F}_k = c'_1 \rightarrow c'_2 \rightarrow \cdots \rightarrow c'_k$. For $k > 1$, the reasoning is considered higher-order ToM \citep{wimmer1983beliefs}.
Psychological studies show that most neurotypical adults can reliably manage second- and third-order ToM \citep{perner1985john}, but performance drops sharply at fourth-order and beyond. Only a minority --- typically those with advanced education or explicit training --- can process such deeply nested beliefs, and even then their responses are slower and less accurate \citep{kinderman1998theory,dunbar2003social}.
Thus, in humans, ToM reasoning with $k \geq 4$ appears to depend on advanced cognitive skills rather than generalizable mechanisms. Accordingly, this study focuses on reasoning orders $k \leq 3$.

\paragraph{Summary}
\

Overall, the complexity of a Sally-Anne task can be controlled through four variables. The scene is mapped to a belief graph $\mathcal{G}$, determined by $\{(n, m), q \mid \mathcal{L}\}$, while the query is mapped to a belief flow $\mathcal{F}$, determined by $k$. The characteristics of these variables are as follows:
\begin{itemize}[leftmargin=0.5cm]
    \item $(n, m)$ denotes the structure of the belief graph $\mathcal{G}$, with $n$ representing the number of \characters ~(nodes) and $m$ the number of \cc ~belief links (edges).
    \item $q \in \mathcal{Q}$ indicates the number of \containers ~involved in the scene.
    \item $k \in \mathcal{K}$ denotes the \order ~of reasoning required to answer the query.
\end{itemize}
Therefore, a family of independently and identically distributed ({\it i.i.d.}) Sally-Anne tasks can be parameterized by the attribute tuple $\{(n,m), q, k \mid \mathcal{L}\}$. Based on these attributes, tasks of different complexities are associated with corresponding data distributions $\mathbb{P}_{\{(n,m),q,k\mid\mathcal{L}\}}$, yielding a collection of {\it i.i.d.} datasets $\mathcal{D}_{\{(n,m),q,k \mid \mathcal{L}\}}$:
\begin{small}
\begin{equation}
\left\{ \big((\text{Scene}, \text{Query}), \text{Answer}\big) \sim \mathbb{P}_{\{(n,m),q,k \mid \mathcal{L}\}} \middle| (n,m)\in\mathcal{S}, q\in\mathcal{Q}, k\in\mathcal{K}, \mathcal{L}=\mathcal{L}_0 \cap \mathcal{L_C} \cap \mathcal{L_B} \cap \mathcal{L_A} \right\}.
\end{equation}
\end{small}

Here, $\mathcal{S}$, $\mathcal{Q}$, and $\mathcal{K}$ denote scalar spaces controlling the complexities of the scene, the query, and the reasoning order, respectively. The language space $\mathcal{L}$ is a predefined semantic pool from which all vocabulary in the task descriptions is drawn, ensuring no out-of-vocabulary terms and thus preserving experimental validity. Specifically, $\mathcal{L}$ is composed of four subspaces: the general narration space $\mathcal{L}_0$, the personalized \character ~semantic pool $\mathcal{L_C}$, the \container ~semantic space $\mathcal{L_B}$, and the \object ~semantic space $\mathcal{L_A}$.

\subsection{Textual Concretization}
\label{sec:data}

\begin{figure}[t]
  \centering
  \includegraphics[width=1\linewidth]{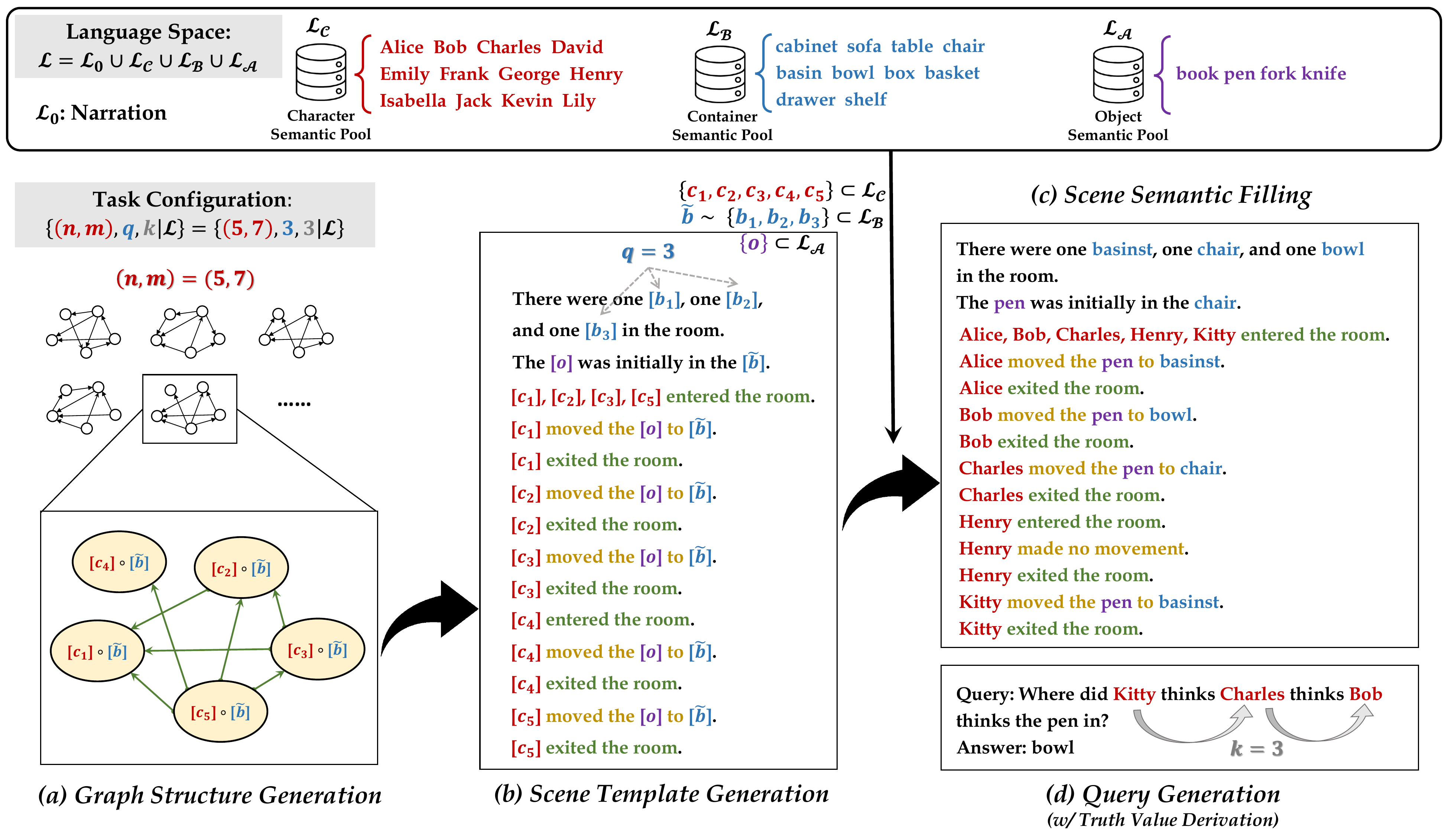}
  \caption{{\bf The pipeline of the textual concretization process under the configuration $\{(n,m), q, k \mid \mathcal{L}\}$.}
  \textbf{\textcolor{red}{(Stage 1)}} We first screen all valid belief graph structures containing $n$ nodes and $m$ directed edges.
  \textbf{\textcolor{red}{(Stage 2)}} For each graph structure (need to double-check its validity), we derive the entrance-exit order of \characters, {\it i.e.}, the \cc ~interaction order. Based on this, we construct a scene template in natural language from $\mathcal{L}_0$, embedding static element placeholders—$n$ \characters, one \object, $q$ \containers, and $n$ \coc ~interactions.
  \textbf{\textcolor{red}{(Stage 3)}} For each scene template, we perform semantic filling of all element placeholders from $\mathcal{L_C}, \mathcal{L_B}, \mathcal{L_A}$, to form the final ``\texttt{Scene}''. By replacing them with different semantics, a single template can be expanded into diverse scenes.
  \textbf{\textcolor{red}{(Stage 4)}} Finally, for each ``\texttt{Scene}'', we sample a belief flow involving $k$ \characters ~to form a ``\texttt{Query}'', and derive the corresponding ground-truth ``\texttt{Answer}'' through designed algorithms, thereby constructing a complete data sample $((\texttt{Scene}, \texttt{Query}), \texttt{Answer})$.}
  \label{fig:data}
\vspace{-0.05in}
\end{figure}

After establishing the graphical abstraction and independent data distributions, each dataset $\mathcal{D}_{\{(n,m),q,k \mid \mathcal{L}\}}$ need to be systematically constructed from the fixed attribute $\{(n,m), q, k \mid \mathcal{L}\}$.
In this process, we need to start from the graph structures with controllable complexity and reconstruct them into the textual form of the Sally-Anne task; thus, it is a process of ``textual concretization''.
This process is illustrated in Figure \ref{fig:data}.

\paragraph{Stage 1: Graph Structure Screening}
\

We first consider screening all possible graph structures with $n$ nodes and $m$ directed edges, to expand the task structure variety (Figure \ref{fig:data}(a)).
Valid structures must satisfy two conditions according to the task characteristics:
\textbf{\textcolor{blue}{Acyclicity:} The graph contains no directed cycle.} Since \characters ~never re-enter once they exit, cyclic graphs cannot correspond to valid textual scenes;
\textbf{\textcolor{blue}{Connectivity:} The graph contains no isolated nodes.} Since every \character ~must interact with at least one other \character, graphs containing isolated nodes are considered invalid, so each node must appear in at least one edge.

In implementation, we first enumerate all permutations of $[1,2,\ldots,n]$, representing $n$ different \characters, each defining a topological order $t_1 \prec t_2 \prec \cdots \prec t_n$ of the graph.\footnote{Even isomorphic graphs are treated as distinct if their topological orders differ, since different index assignments yield semantically different textual scenes. Thus, enumerating all valid topological orders is necessary.}
To ensure the acyclicity, edges are then generated only from later to earlier nodes $(t_j, t_i), j > i$. For each order, there are $\binom{n}{2}$ candidate edges, from which we choose $m$ to form the edge set $\mathcal{E}$, yielding $\binom{\binom{n}{2}}{m}$ possible combinations.
Each candidate is retained only if it has no isolated nodes. Finally, for every valid graph, we assign two placeholders to each node --- \characters ~and \containers --- representing the \coc ~interaction when a \character ~exits the scene, {\it i.e.},  $\mathcal{V} = \{\textsf{[c}_{t_i}\textsf{]} \circ \textsf{[}\tilde{\textsf{b}}\textsf{]} \}_{i=1}^n$.
The pseudo code of this screening algorithm is presented in Algorithm \ref{alg:1}.


\paragraph{Stage 2: Scene Template Mapping}
\

Each graph structure inherently encodes the \cc ~interaction, namely the entrance-exit order of the $n$ \characters. At this stage, we expand every valid graph structure into a textual representation, using natural language to describe the complete interaction event and thereby constructing a scene template containing static element placeholders (Figure \ref{fig:data}(b)).

First, we derive the entrance–exit order of the $n$ \characters. In Algorithm \ref{alg:1}, all valid graph structures and their associated topological orders are enumerated, where each order specifies the sequence of character exits. To achieve this, we maintain a dynamic list of \characters ~currently present in the scene and traverse the topological order from left to right.
We initialize an empty ``room'' list to track the 
\characters ~in the scene. At each iteration, if the current node is already in the ``room'', it exits; otherwise, it enters with all of its parent nodes, then exits while its parents remain. Afterward, we verify that every node in the room is indeed a parent of the current node; if not, the structure is invalid and discarded.
Violating this condition implies an interaction between two nodes without a corresponding edge in the belief graph, contradicting its definition that a parent must be present when the child exits to enable their interaction. We term this dynamic validity check our third condition: \textbf{\textcolor{blue}{Room-induced Closure}}.
The pseudo code of this sub-stage is presented in Algorithm \ref{alg:2}.

After obtaining these character orders, we expand them into a coherent narrative by linking these placeholders with textual descriptions, thereby producing a scene template.
We initialize the \object ~$\textsf{[a]}$ ~and the $q$ \containers ~$\{\textsf{[b}_i\textsf{]}\}_{i=1}^q$ in the scene with placeholders, then overlay narration semantics onto the character order to form coherent expressions.
The pseudo code of this sub-stage is presented in Algorithm \ref{alg:3}.

To make this stage more transparent, Figure \ref{fig:mapping-case57} provides some examples of the two serial sub-stages under one configuration.

\paragraph{Stage 3: Scene Generation}
\label{sec:ssf}
\ 

After generating scene templates over all valid graph structures, we fill all placeholders in the scene template with semantic items from the language space $\mathcal{L}$, to form the final natural language description of the scene (Figure \ref{fig:data}(c)).
Specifically, $n$ \characters ~$\{\textsf{[c}_i\textsf{]}\}_{i=1}^n$ are sampled {\it without replacement} from $\mathcal{L_C}$, $q$ \containers ~$\{\textsf{[b}_i\textsf{]}\}_{i=1}^q$ are sampled {\it without replacement} from $\mathcal{L_B}$, and a single \object ~$\textsf{[a]}$ is sampled from $\mathcal{L_A}$. Each node attribute $\textsf{[}\tilde{\textsf{b}}\textsf{]}$, representing \coc ~interaction, is then sampled {\it with replacement} from $\{\textsf{[b}_i\textsf{]}\}_{i=1}^q$. Sampling in this stage is critical to expanding semantic diversity.

In implementation, we partition the language spaces $\mathcal{L_C}$, $\mathcal{L_B}$, and $\mathcal{L_A}$ into disjoint subsets of size $n$, $q$, and $|\mathcal{L_A}|$, respectively. This design stems from our graph screening step, where we treat isomorphic graphs with different topological orders as distinct. To support this, each node must be assigned unique attributes, ensuring that structurally isomorphic graphs still yield different scene descriptions. Disjoint partitioning thus guarantees that the semantics assigned to each \character ~and \container ~placeholder ($\textsf{[c}_i\textsf{]}$, $\textsf{[b}_i\textsf{]}$) are unique and cannot overlap with those of any other $\textsf{[c}_j\textsf{]}$ or $\textsf{[b}_j\textsf{]}$ $(j \neq i)$.
After replacing the placeholders $\{\textsf{[c}_i\textsf{]}\}_{i=1}^n$, $\{\textsf{[b}_i\textsf{]}\}_{i=1}^q$, and $\textsf{[a]}$ with natural language into $\{c_i\}_{i=1}^n$, $\{b_i\}_{i=1}^q$, and $\{a\}$, we obtain the final ``\texttt{Scene}''.
Figure \ref{fig:semantic-case57} and \ref{fig:semantic-case69} provide detailed examples of this generation and expansion process.

\paragraph{Stage 4: Query Generation and Answer Derivation}
\ 

After obtaining the scene, we need to design a $k$-\order ~query for each scene.
For each query, we sample $k$ \characters ~{\it without replacement} from $\{c_i\}_{i=1}^n$ in order, to form the belief flow, which is subsequently converted into a textual ``\texttt{Query}''.

After constructing the ``\texttt{Scene}'' and ``\texttt{Query}'', one final issue remains before training and testing: how to obtain the ``\texttt{Answer}'' for each query. Therefore, we designed a truth derivation algorithm based on the belief graph $\mathcal{G}$ corresponding to each scene.

Each query corresponds to a belief flow $\mathcal{F}_k = c'_1 \rightarrow c'_2 \rightarrow \cdots \rightarrow c'_k$, whose ground truth is denoted by $T(\mathcal{F}_k)$. Let the true beliefs of $c'_1, c'_2, \ldots, c'_k$ be $b'_1, b'_2, \ldots, b'_k$, respectively.
For $k = 1$, the truth value is simply the first character’s belief, i.e., $T(\mathcal{F}_1) = b'_1$. For $k = 2$, $T(\mathcal{F}_2)$ equals $c'_2$’s actual belief $b'_{2'}$ if the edge $(c'_1, c'_2) \in \mathcal{G}$ exists in the belief graph; otherwise, $c'_1$ defaults to its own belief $b'_{1'}$. This is immediate from the definition of the belief graph edge.
When $k \geq 3$, the number of possible cases grows exponentially. For each pair of nodes, there are three possible edge states: no edge, a forward edge, or a backward edge. Consequently, the total number of cases is $3^{(k(k-1)/2)}$, yielding a complexity of $\mathcal{O}(3^{k^2})$. For $k=3$, this results in 27 distinct cases. We provide the truth value table for all cases in Table \ref{tab:structure} to ensure that every possible case has a corresponding answer.

\begin{table}[t]

\centering
\footnotesize
\renewcommand\arraystretch{1.2}
\caption{Truth derivation cases for queries at $k=3$. The true beliefs of $c'_1, c'_2, c'_3$ are $b'_1, b'_2, b'_3$, respectively. The query template is \textit{``What does $c'_1$ think $c'_2$ thinks $c'_3$ thinks the [\object] is in (which [\container])?''}, yielding the belief flow $c'_1 \rightarrow c'_2 \rightarrow c'_3$. For the belief subgraph defined by these three characters, there are $3^3 = 27$ possible configurations (including 2 cyclic, hence invalid, structures). We enumerate the final query answer for each case.}

\setlength{\tabcolsep}{5mm}{
  \resizebox{1\textwidth}{!}{
\begin{tabular}{c|c|c|c|c|c}

\toprule
Belief Sub-Graph & Answer & Belief Sub-Graph & Answer & Belief Sub-Graph & Answer \\
\midrule

\begin{tikzpicture}[scale=0.8, every node/.style={font=\small}]
    \draw[thick] (-2,0) circle(0.3cm);
    \draw[thick] (0,0) circle(0.3cm);
    \draw[thick] (2,0) circle(0.3cm);
    \node (A) at (-2,0) {$c'_1$};
    \node (B) at (0,0) {$c'_2$};
    \node (C) at (2,0) {$c'_3$};
\end{tikzpicture} & $b'_1$
&

\begin{tikzpicture}[scale=0.8, every node/.style={font=\small}]
    \draw[thick] (-2,0) circle(0.3cm);
    \draw[thick] (0,0) circle(0.3cm);
    \draw[thick] (2,0) circle(0.3cm);
    \node (A) at (-2,0) {$c'_1$};
    \node (B) at (0,0) {$c'_2$};
    \node (C) at (2,0) {$c'_3$};
    \draw[thick][->] (A) to[out=30, in=150] (C);
\end{tikzpicture} & $b'_3$
&

\begin{tikzpicture}[scale=0.8, every node/.style={font=\small}]
    \draw[thick] (-2,0) circle(0.3cm);
    \draw[thick] (0,0) circle(0.3cm);
    \draw[thick] (2,0) circle(0.3cm);
    \node (A) at (-2,0) {$c'_1$};
    \node (B) at (0,0) {$c'_2$};
    \node (C) at (2,0) {$c'_3$};
    \draw[thick][->] (C) to[out=150, in=30] (A);
\end{tikzpicture} & $b'_1$ \\
\midrule

\begin{tikzpicture}[scale=0.8, every node/.style={font=\small}]
    \draw[thick] (-2,0) circle(0.3cm);
    \draw[thick] (0,0) circle(0.3cm);
    \draw[thick] (2,0) circle(0.3cm);
    \node (A) at (-2,0) {$c'_1$};
    \node (B) at (0,0) {$c'_2$};
    \node (C) at (2,0) {$c'_3$};
    \draw[thick][->] (A) -- (B) node[right, left] {};
\end{tikzpicture} & $b'_2$
&

\begin{tikzpicture}[scale=0.8, every node/.style={font=\small}]
    \draw[thick] (-2,0) circle(0.3cm);
    \draw[thick] (0,0) circle(0.3cm);
    \draw[thick] (2,0) circle(0.3cm);
    \node (A) at (-2,0) {$c'_1$};
    \node (B) at (0,0) {$c'_2$};
    \node (C) at (2,0) {$c'_3$};
    \draw[thick][->] (A) -- (B) node[right, left] {};
    \draw[thick][->] (A) to[out=30, in=150] (C);
\end{tikzpicture} & $b'_2$
&

\begin{tikzpicture}[scale=0.8, every node/.style={font=\small}]
    \draw[thick] (-2,0) circle(0.3cm);
    \draw[thick] (0,0) circle(0.3cm);
    \draw[thick] (2,0) circle(0.3cm);
    \node (A) at (-2,0) {$c'_1$};
    \node (B) at (0,0) {$c'_2$};
    \node (C) at (2,0) {$c'_3$};
    \draw[thick][->] (A) -- (B) node[right, left] {};
    \draw[thick][->] (C) to[out=150, in=30] (A);
\end{tikzpicture} & $b'_2$ \\
\midrule

\begin{tikzpicture}[scale=0.8, every node/.style={font=\small}]
    \draw[thick] (-2,0) circle(0.3cm);
    \draw[thick] (0,0) circle(0.3cm);
    \draw[thick] (2,0) circle(0.3cm);
    \node (A) at (-2,0) {$c'_1$};
    \node (B) at (0,0) {$c'_2$};
    \node (C) at (2,0) {$c'_3$};
    \draw[thick][->] (B) -- (A) node[left, right] {};
\end{tikzpicture} & $b'_1$
&
\begin{tikzpicture}[scale=0.8, every node/.style={font=\small}]
    \draw[thick] (-2,0) circle(0.3cm);
    \draw[thick] (0,0) circle(0.3cm);
    \draw[thick] (2,0) circle(0.3cm);
    \node (A) at (-2,0) {$c'_1$};
    \node (B) at (0,0) {$c'_2$};
    \node (C) at (2,0) {$c'_3$};
    \draw[thick][->] (B) -- (A) node[left, right] {};
    \draw[thick][->] (A) to[out=30, in=150] (C);
\end{tikzpicture} & $b'_3$ 
&

\begin{tikzpicture}[scale=0.8, every node/.style={font=\small}]
    \draw[thick] (-2,0) circle(0.3cm);
    \draw[thick] (0,0) circle(0.3cm);
    \draw[thick] (2,0) circle(0.3cm);
    \node (A) at (-2,0) {$c'_1$};
    \node (B) at (0,0) {$c'_2$};
    \node (C) at (2,0) {$c'_3$};
    \draw[thick][->] (B) -- (A) node[left, right] {};
    \draw[thick][->] (C) to[out=150, in=30] (A);
\end{tikzpicture} & $b'_1$ \\
\midrule

\begin{tikzpicture}[scale=0.8, every node/.style={font=\small}]
    \draw[thick] (-2,0) circle(0.3cm);
    \draw[thick] (0,0) circle(0.3cm);
    \draw[thick] (2,0) circle(0.3cm);
    \node (A) at (-2,0) {$c'_1$};
    \node (B) at (0,0) {$c'_2$};
    \node (C) at (2,0) {$c'_3$};
    \draw[thick][->] (B) -- (C) node[right, left] {};
\end{tikzpicture} & $b'_1$
&

\begin{tikzpicture}[scale=0.8, every node/.style={font=\small}]
    \draw[thick] (-2,0) circle(0.3cm);
    \draw[thick] (0,0) circle(0.3cm);
    \draw[thick] (2,0) circle(0.3cm);
    \node (A) at (-2,0) {$c'_1$};
    \node (B) at (0,0) {$c'_2$};
    \node (C) at (2,0) {$c'_3$};
    \draw[thick][->] (B) -- (C) node[right, left] {};
    \draw[thick][->] (A) to[out=30, in=150] (C);
\end{tikzpicture} & $b'_3$
&

\begin{tikzpicture}[scale=0.8, every node/.style={font=\small}]
    \draw[thick] (-2,0) circle(0.3cm);
    \draw[thick] (0,0) circle(0.3cm);
    \draw[thick] (2,0) circle(0.3cm);
    \node (A) at (-2,0) {$c'_1$};
    \node (B) at (0,0) {$c'_2$};
    \node (C) at (2,0) {$c'_3$};
    \draw[thick][->] (B) -- (C) node[right, left] {};
    \draw[thick][->] (C) to[out=150, in=30] (A);
\end{tikzpicture} & $b'_1$ \\
\midrule

\begin{tikzpicture}[scale=0.8, every node/.style={font=\small}]
    \draw[thick] (-2,0) circle(0.3cm);
    \draw[thick] (0,0) circle(0.3cm);
    \draw[thick] (2,0) circle(0.3cm);
    \node (A) at (-2,0) {$c'_1$};
    \node (B) at (0,0) {$c'_2$};
    \node (C) at (2,0) {$c'_3$};
    \draw[thick][->] (A) -- (B) node[right, left] {};
    \draw[thick][->] (B) -- (C) node[right, left] {};
\end{tikzpicture} & $b'_2$ 
&

\begin{tikzpicture}[scale=0.8, every node/.style={font=\small}]
    \draw[thick] (-2,0) circle(0.3cm);
    \draw[thick] (0,0) circle(0.3cm);
    \draw[thick] (2,0) circle(0.3cm);
    \node (A) at (-2,0) {$c'_1$};
    \node (B) at (0,0) {$c'_2$};
    \node (C) at (2,0) {$c'_3$};
    \draw[thick][->] (A) -- (B) node[right, left] {};
    \draw[thick][->] (B) -- (C) node[right, left] {};
    \draw[thick][->] (A) to[out=30, in=150] (C);
\end{tikzpicture} & $b'_3$ 
&

\begin{tikzpicture}[scale=0.8, every node/.style={font=\small}]
    \draw[thick] (-2,0) circle(0.3cm);
    \draw[thick] (0,0) circle(0.3cm);
    \draw[thick] (2,0) circle(0.3cm);
    \node (A) at (-2,0) {$c'_1$};
    \node (B) at (0,0) {$c'_2$};
    \node (C) at (2,0) {$c'_3$};
    \draw[thick][->] (A) -- (B) node[right, left] {};
    \draw[thick][->] (B) -- (C) node[right, left] {};
    \draw[thick][->] (C) to[out=150, in=30] (A);
\end{tikzpicture} & cycle, invalid \\
\midrule

\begin{tikzpicture}[scale=0.8, every node/.style={font=\small}]
    \draw[thick] (-2,0) circle(0.3cm);
    \draw[thick] (0,0) circle(0.3cm);
    \draw[thick] (2,0) circle(0.3cm);
    \node (A) at (-2,0) {$c'_1$};
    \node (B) at (0,0) {$c'_2$};
    \node (C) at (2,0) {$c'_3$};
    \draw[thick][->] (B) -- (A) node[left, right] {};
    \draw[thick][->] (B) -- (C) node[right, left] {};
\end{tikzpicture} & $b'_1$
&

\begin{tikzpicture}[scale=0.8, every node/.style={font=\small}]
    \draw[thick] (-2,0) circle(0.3cm);
    \draw[thick] (0,0) circle(0.3cm);
    \draw[thick] (2,0) circle(0.3cm);
    \node (A) at (-2,0) {$c'_1$};
    \node (B) at (0,0) {$c'_2$};
    \node (C) at (2,0) {$c'_3$};
    \draw[thick][->] (B) -- (A) node[left, right] {};
    \draw[thick][->] (B) -- (C) node[right, left] {};
    \draw[thick][->] (A) to[out=30, in=150] (C);
\end{tikzpicture} & $b'_3$ 
&

\begin{tikzpicture}[scale=0.8, every node/.style={font=\small}]
    \draw[thick] (-2,0) circle(0.3cm);
    \draw[thick] (0,0) circle(0.3cm);
    \draw[thick] (2,0) circle(0.3cm);
    \node (A) at (-2,0) {$c'_1$};
    \node (B) at (0,0) {$c'_2$};
    \node (C) at (2,0) {$c'_3$};
    \draw[thick][->] (B) -- (A) node[left, right] {};
    \draw[thick][->] (B) -- (C) node[right, left] {};
    \draw[thick][->] (C) to[out=150, in=30] (A);
\end{tikzpicture} & $b'_1$ \\
\midrule

\begin{tikzpicture}[scale=0.8, every node/.style={font=\small}]
    \draw[thick] (-2,0) circle(0.3cm);
    \draw[thick] (0,0) circle(0.3cm);
    \draw[thick] (2,0) circle(0.3cm);
    \node (A) at (-2,0) {$c'_1$};
    \node (B) at (0,0) {$c'_2$};
    \node (C) at (2,0) {$c'_3$};
    \draw[thick][->] (C) -- (B) node[left, right] {};
\end{tikzpicture} & $b'_1$
&

\begin{tikzpicture}[scale=0.8, every node/.style={font=\small}]
    \draw[thick] (-2,0) circle(0.3cm);
    \draw[thick] (0,0) circle(0.3cm);
    \draw[thick] (2,0) circle(0.3cm);
    \node (A) at (-2,0) {$c_1$};
    \node (B) at (0,0) {$c_2$};
    \node (C) at (2,0) {$c_3$};
    \draw[thick][->] (C) -- (B) node[left, right] {};
    \draw[thick][->] (A) to[out=30, in=150] (C);
\end{tikzpicture} & $b'_3$
&

\begin{tikzpicture}[scale=0.8, every node/.style={font=\small}]
    \draw[thick] (-2,0) circle(0.3cm);
    \draw[thick] (0,0) circle(0.3cm);
    \draw[thick] (2,0) circle(0.3cm);
    \node (A) at (-2,0) {$c'_1$};
    \node (B) at (0,0) {$c'_2$};
    \node (C) at (2,0) {$c'_3$};
    \draw[thick][->] (C) -- (B) node[left, right] {};
    \draw[thick][->] (C) to[out=150, in=30] (A);
\end{tikzpicture} & $b'_1$ \\
\midrule

\begin{tikzpicture}[scale=0.8, every node/.style={font=\small}]
    \draw[thick] (-2,0) circle(0.3cm);
    \draw[thick] (0,0) circle(0.3cm);
    \draw[thick] (2,0) circle(0.3cm);
    \node (A) at (-2,0) {$c'_1$};
    \node (B) at (0,0) {$c'_2$};
    \node (C) at (2,0) {$c'_3$};
    \draw[thick][->] (A) -- (B) node[right, left] {};
    \draw[thick][->] (C) -- (B) node[left, right] {};
\end{tikzpicture} & $b'_2$
&

\begin{tikzpicture}[scale=0.8, every node/.style={font=\small}]
    \draw[thick] (-2,0) circle(0.3cm);
    \draw[thick] (0,0) circle(0.3cm);
    \draw[thick] (2,0) circle(0.3cm);
    \node (A) at (-2,0) {$c'_1$};
    \node (B) at (0,0) {$c'_2$};
    \node (C) at (2,0) {$c'_3$};
    \draw[thick][->] (A) -- (B) node[right, left] {};
    \draw[thick][->] (C) -- (B) node[left, right] {};
    \draw[thick][->] (A) to[out=30, in=150] (C);
\end{tikzpicture} & $b'_2$
&

\begin{tikzpicture}[scale=0.8, every node/.style={font=\small}]
    \draw[thick] (-2,0) circle(0.3cm);
    \draw[thick] (0,0) circle(0.3cm);
    \draw[thick] (2,0) circle(0.3cm);
    \node (A) at (-2,0) {$c'_1$};
    \node (B) at (0,0) {$c'_2$};
    \node (C) at (2,0) {$c'_3$};
    \draw[thick][->] (A) -- (B) node[right, left] {};
    \draw[thick][->] (C) -- (B) node[left, right] {};
    \draw[thick][->] (C) to[out=150, in=30] (A);
\end{tikzpicture} & $b'_2$ \\
\midrule

\begin{tikzpicture}[scale=0.8, every node/.style={font=\small}]
    \draw[thick] (-2,0) circle(0.3cm);
    \draw[thick] (0,0) circle(0.3cm);
    \draw[thick] (2,0) circle(0.3cm);
    \node (A) at (-2,0) {$c'_1$};
    \node (B) at (0,0) {$c'_2$};
    \node (C) at (2,0) {$c'_3$};
    \draw[thick][->] (B) -- (A) node[left, right] {};
    \draw[thick][->] (C) -- (B) node[left, right] {};
\end{tikzpicture} & $b'_1$
&

\begin{tikzpicture}[scale=0.8, every node/.style={font=\small}]
    \draw[thick] (-2,0) circle(0.3cm);
    \draw[thick] (0,0) circle(0.3cm);
    \draw[thick] (2,0) circle(0.3cm);
    \node (A) at (-2,0) {$c'_1$};
    \node (B) at (0,0) {$c'_2$};
    \node (C) at (2,0) {$c'_3$};
    \draw[thick][->] (B) -- (A) node[left, right] {};
    \draw[thick][->] (C) -- (B) node[left, right] {};
    \draw[thick][->] (A) to[out=30, in=150] (C);
\end{tikzpicture} & cycle, invalid
&

\begin{tikzpicture}[scale=0.8, every node/.style={font=\small}]
    \draw[thick] (-2,0) circle(0.3cm);
    \draw[thick] (0,0) circle(0.3cm);
    \draw[thick] (2,0) circle(0.3cm);
    \node (A) at (-2,0) {$c'_1$};
    \node (B) at (0,0) {$c'_2$};
    \node (C) at (2,0) {$c'_3$};
    \draw[thick][->] (B) -- (A) node[left, right] {};
    \draw[thick][->] (C) -- (B) node[left, right] {};
    \draw[thick][->] (C) to[out=150, in=30] (A);
\end{tikzpicture} & $b'_1$ \\
\midrule

\end{tabular}
}
}

\label{tab:structure}%

\end{table}

\subsection{Evaluation Protocol}
\label{sec:protocol}

High-order ToM generalization is referred to as the ability of a model that has mastered reasoning at a small depth $k$ to extrapolate to previously unseen larger depths $k$.
This transition is interpreted as a qualitative leap in core ToM competence.
Formally, for any fixed complexity variable $(n,m)$ and $q$, we partition the data set distribution $\mathbb{P}_{\{(n,m),q,k \mid \mathcal{L}\}}$ into training set $\mathcal{D}_{l}$ for learning and test set $\mathcal{D}_{g}$ for generalization:
\begin{equation}
    \begin{aligned}
        &\mathcal{D}_{l} = \{D \sim \mathbb{P}_{((n, m), q, k' \mid \mathcal{L})} \mid k' \in \mathcal{K}_{l}\},
        ~~\mathcal{D}_{g} = \{D \sim \mathbb{P}_{((n, m), q, k' \mid \mathcal{L})} \mid k' \in \mathcal{K}_{g}\}, \\
        &\quad \text{where} \quad
        \mathcal{K}_{g} \cup \mathcal{K}_{l} = \mathcal{K},~ \mathcal{K}_{g} \cap \mathcal{K}_{l} = \emptyset,~  \bm{\min(\mathcal{K}_{g}) > \max(\mathcal{K}_{l})}.
    \end{aligned}
\end{equation}
The generalization process in general can be denoted as $\mathcal{D}_{l} \xrightarrow{k \uparrow} \mathcal{D}_{g}$.

We varied task complexity to construct parallel experimental groups, ensuring the robustness of our evaluation. Specifically, we set $n \in \{4,5,6\}$ to cover three node scales and then specified the number of edges $m$. Our objective was to maintain a moderate graph density under each $(n,m)$ configuration, thereby avoiding cases where overly sparse graphs yielded insufficient character interactions, while overly dense graphs led to information overload.
Specifically, following \citet{easley2010networks}, we define the graph density as
\begin{equation}
\text{density}(G) = \frac{m}{\binom{n}{2}} = \frac{2m}{n(n-1)}.
\end{equation}
and restrict the candidate $(n,m)$ configurations to the medium-density range $\text{density}(G) \in [0.40,0.80]$.
The final selected configurations are:
when $n=4$, $m=4$ (density $0.67$);
when $n=5$, $m=7,8$ (densities $0.70,0.80$);
when $n=6$, $m=7,8,9$ (densities $0.47,0.53,0.60$).
These densities are evenly distributed within the “neither too sparse nor too dense” interval.
In addition, $q$ determines the search space of the final options: if a scene contains $q$ \containers, the random accuracy is $1/q$. We set $q \in {3,4}$ to vary this baseline.
Overall, we set the spaces of all complexity variables as follows:
\begin{equation}
    \begin{aligned}
        (n,m) \in \mathcal{S} = \{(4,4), (5,7), (5,8), (6,7), (6,8), (6,9)\}, ~~~~q \in \mathcal{Q} = \{3,4\} \\
    \end{aligned}
\end{equation}

Based on this setting, we form {\bf 12 parallel experimental groups}, each corresponding to a unique configuration of $((n,m), q) \in \mathcal{S} \times \mathcal{Q}$.
Then we let $\mathcal{K} = \{1,2,3\}$, where $\mathcal{K}_l = \{1\}$, $\mathcal{K}_g = \{2,3\}$, so within each group, our ToMNN was trained exclusively on $\mathcal{D}_l$ with $k=1$ (first-order ToM) and subsequently tested on $\mathcal{D}_g$ with $k = 2,3$ (second- and third-order ToM).
Regarding the dataset size under each group, the statistics are provided in Table \ref{tab:datasize}, with further explanation in Section \ref{sec:data-scale}.



\begin{table}[t]
\centering
\footnotesize
\renewcommand\arraystretch{1.2}
\caption{Dataset sizes under different configurations $(n,m,q)$.}
\setlength{\tabcolsep}{2mm}{
\resizebox{1\textwidth}{!}{
\begin{tabular}{c|c|c|c|c|c}
\toprule
{\bf Configuration} & {\bf Training Size} & {\bf Testing Size} 
& {\bf Configuration} & {\bf Training Size} & {\bf Testing Size} \\
\midrule

(4,4,3) & \multirow{2}{*}{1,306,368} & \multirow{2}{*}{93,312} 
& (6,7,3) & \multirow{6}{*}{1,032,192} & \multirow{6}{*}{73,728} \\
(4,4,4) & & & (6,7,4) & & \\
\cline{1-3}

(5,7,3) & \multirow{4}{*}{1,161,216} & \multirow{4}{*}{82,944} 
& (6,8,3) & & \\
(5,8,3) & & & (6,8,4) & & \\
(5,7,4) & & & (6,9,3) & & \\
(5,8,4) & & & (6,9,4) & & \\
\bottomrule
\end{tabular}}}
\label{tab:datasize}%
\end{table}

\subsection{Data Scale of $\mathcal{D}_{\{(n,m),q,k \mid \mathcal{L}\}}$}
\label{sec:data-scale}

In constructing the $\mathcal{D}_{\{(n,m),q,k \mid \mathcal{L}\}}$, we began with the initial variables $\{(n,m),q,k \mid \mathcal{L}\}$ and two stages involved diversification: (1) {\it graph structure screening}, which ensured diversity in interaction topological structures, and (2) {\it scene generation}, which introduced diversity in semantic representations. Based on these, we defined and calculated the dataset size under all variable configurations $\{(n,m),q,k \mid \mathcal{L}\}$.

\paragraph{Topological Structure Expansion}
\

During the {\it graph structure screening} stage (Section \ref{sec:data}, Stage 1), all structures with $n$ labeled nodes (each assigned a unique attribute) and $m$ directed edges are required to satisfy the following validity conditions:

\begin{enumerate}[leftmargin=0.9cm]
    \item \textbf{\textcolor{blue}{Acyclicity:}} The graph contains no directed cycle.
    \item \textbf{\textcolor{blue}{Connectivity:}} The graph contains no isolated nodes, {\it i.e.}, each node must appear in at least one edge.
\end{enumerate}

First, for the \textbf{Acyclicity}, we enforce this by fixing a canonical topological order $1 \prec 2 \prec \cdots \prec n$ and only allowing edges to point backward in that order. The admissible edge set is
\begin{equation}
E^\star=\{(j,i):1 \preceq i \prec j \preceq n\}, \qquad M=|E^\star|=\binom{n}{2}.
\end{equation}
Each candidate graph is a subset $E\subseteq E^\star$ with $|E|=m$.
Under this topological order, and in accordance with the task-specific requirements (Section \ref{sec:data}, Stage 2) during the {\it scene template generation} stage, a valid graph must further satisfy the following condition:

\begin{enumerate}[leftmargin=0.9cm]
    \setcounter{enumi}{2}
    \item \textbf{\textcolor{blue}{Room-induced Closure:}} 
    Let $S_v$ denote the set of nodes currently present in the ``room'' immediately before processing node $v$.
    The iteration begins with an empty room, $S_1 = \varnothing$. 
    At each step $v=1,\dots,n$, the following rules apply:
    \begin{align}
        &\text{(Closure constraint)} \quad S_v \setminus \{v\} \subseteq P(v), \\
        &\text{(Room update)} \quad S_{v+1} = (S_v \cup P(v)) \setminus \{v\},
    \end{align}
    where $P(v) = \{\,u : (u,v)\in E\,\}$ denotes the parent set of $v$. 
    In words, every individual already in the room (except $v$ itself) must also be a parent of $v$, and after $v$ is processed, the room is updated by admitting all of $v$’s parents and removing $v$ itself.
\end{enumerate}

With the above constraints, the room state evolves deterministically.  
Since $S_v \setminus \{v\} \subseteq P(v)$, we have
\begin{equation}
S_{v+1} = (S_v \cup P(v)) \setminus \{v\} = P(v).
\end{equation}
Let $r_v = |P(v)|$ and $s_v = |S_v|$. This yields the recurrence
\begin{equation}
s_1 = 0, \qquad s_{v+1} = r_v, \qquad s_v \leq r_v \leq |V_{\succ v}| \quad (\forall v),
\end{equation}
where $V_{\succ v} := \{\, u \in V : v \prec u \,\}$ is the set of nodes that follow $v$ in the topological order, so $|V_{\succ v}| = n-v$ under the canonical labeling.

At each step $v$, among the $|V_{\succ v}|$ potential parents of $v$, exactly $s_v$ are already present in the room and must be included; the remaining $r_v - s_v$ parents can be freely selected from the $|V_{\succ v}| - s_v$ optional candidates.  
Thus, the number of admissible choices for $P(v)$ is
\begin{equation}
\binom{|V_{\succ v}| - s_v}{\,r_v - s_v\,}.
\end{equation}

Summing over all feasible parent-size sequences $\{r_v\}$ with $\sum_{v=1}^n r_v = m$, we obtain the count of valid structures under a fixed topological order (before enforcing connectivity):
\begin{equation}
N_0'(n,m)=
\sum_{\substack{r_1,\dots,r_n \\
s_1=0,\ s_{v+1}=r_v \\
s_v \leq r_v \leq |V_{\succ v}| \\
\sum r_v = m}}
\ \prod_{v=1}^{n}\binom{|V_{\succ v}| - s_v}{\,r_v - s_v\,}.
\end{equation}

Finally, to eliminate isolated nodes, we apply an inclusion–exclusion principle over all subsets $X \subseteq V=\{1,\dots,n\}$.  
For $X$, enforcing that every $u \in X$ is isolated sets $r_u=0$ and removes $u$ from all later candidate pools.  
This shrinks the optional pool at step $v$ from $|V_{\succ v}| - s_v$ to $|V_{\succ v}| - s_v - |X_{\succ v}|$, where $X_{\succ v} := X \cap V_{\succ v} = \{\, u \in X : v \prec u \,\}$.  
The resulting count is
\begin{equation}
N'(n,m)=
\sum_{X\subseteq V}(-1)^{|X|}
\sum_{\substack{r_1,\dots,r_n \\
s_1=0,\ s_{v+1}=r_v \\
s_v \leq r_v \leq |V_{\succ v}| - |X_{\succ v}| \\
r_u=0,\ \forall u\in X \\
\sum r_v = m}}
\ \prod_{v=1}^{n}\binom{|V_{\succ v}| - s_v - |X_{\succ v}|}{\,r_v - s_v\,}.
\end{equation}

Because isomorphic graphs under different topological orders are treated as distinct, the final count is
\begin{equation}
N(n,m) = n! \, N'(n,m).
\label{eq:Nfinal}
\end{equation}

Back to our experimental setup, we consider six $(n,m)$ configurations: $(4,4)$, $(5,7)$, $(5,8)$, $(6,7)$, $(6,8)$, and $(6,9)$. Substituting these into the counting procedure described above yields the exact number of valid graph structures for each configuration, as shown in Table~\ref{tab:number}. We observe that the total number of structures increases substantially with larger $n$. To maintain consistency across all configurations, we adopt the minimum count, $N(4,4)=120$, as the reference. Accordingly, {\bf we restrict the dataset to 120 distinct structures (112 for training and 8 held out for testing).}

\begin{table}[t]
\centering
\caption{Number of valid DAG structures $N'(n,m)$ under a fixed topological order, and the final count $N(n,m)=n!\,N'(n,m)$ when all topological orders are distinguished.}
\setlength{\tabcolsep}{10pt}
\renewcommand{\arraystretch}{1.2}
\begin{tabular}{c c c c}
\toprule
$n$ & $m$ & $N'(n,m)$ & $N(n,m)=n!\,N'(n,m)$ \\
\midrule
4 & 4 & $5$ & \underline{$120$} \\
5 & 7 & $15$ & $1{,}800$ \\
5 & 8 & $9$ & $1{,}080$ \\
6 & 7 & $71$ & $51{,}120$ \\
6 & 8 & $83$ & $59{,}760$ \\
6 & 9 & $82$ & $59{,}040$ \\
\bottomrule
\end{tabular}
\label{tab:number}
\end{table}

\paragraph{Semantics Expansion}
\

During the mapping from graph structures to textual scenes and queries, we conducted multiple rounds of semantic sampling to generate large-scale and diverse data samples. Consequently, a single graph structure can correspond to numerous distinct textual scenes and queries.
Specifically, the sizes of the semantic spaces predefined by us are $|\mathcal{L_C}| = 12$, $|\mathcal{L_B}| = 10$, and $|\mathcal{L_A}| = 4$, with the detailed semantic pool shown at the top of Figure \ref{fig:data}.
For \character ~initialization, $\mathcal{L_C}$ is partitioned into $n$ equal parts, and each of $n$ \character ~selects semantics from its assigned sub-interval, yielding 
\begin{equation}
    \#_{\mathcal{C}}(n) = \prod_{i=1}^n (\lfloor \tfrac{12}{n} i \rfloor - \lfloor \tfrac{12}{n}(i-1) \rfloor)
\end{equation}
possible combinations.
For \container ~initialization, $\mathcal{L_B}$ is divided into $q$ equal parts, and each of $q$ container selects semantics from its sub-interval, yielding
\begin{equation}
    \#_{\mathcal{B}}(q) = \prod_{i=1}^q (\lfloor \tfrac{10}{q} i \rfloor - \lfloor \tfrac{10}{q}(i-1) \rfloor)    
\end{equation}
possible combinations.
For \object ~initialization, since only one \object ~is selected, there are $C_4^1$ possible combinations.
Finally, node attributes are sampled with replacement from $\mathcal{B}$, yielding a single random combination.
Thus, the total number of scenes generated from a single graph structure under the $\{(n, m), q, k \mid \mathcal{L}\}$ is:
\begin{equation}
    \#_{\mathcal{C}}(n) * \#_{\mathcal{B}}(q) * C_4^1
\label{eq:size}
\end{equation}

\paragraph{Summary}
\

During the textual concretization process, we start from $\{(n, m), q, k \mid \mathcal{L}\}$ and first expand $N(n,m)$ graph structures (Eq.\ref{eq:Nfinal}), each corresponding to a unique scene template. Each scene template is then expanded into $4 * \#_{\mathcal{C}}(n) * \#_{\mathcal{B}}(q)$ scenes (Eq.\ref{eq:size}), and each scene is paired with a randomly sampled $k$-order query along with its derived ground truth.
Consequently, the data scale of $\mathcal{D}_{\{(n,m),q,k \mid \mathcal{L}\}}$ is given by:
\begin{equation}
    \left| \mathcal{D}_{\{(n,m),q,k \mid \mathcal{L}\}} \right| = 4 * N(n,m) * \#_{\mathcal{C}}(n) * \#_{\mathcal{B}}(q)
\end{equation}
The results calculated from this formula are reported in Table \ref{tab:datasize}. Since $\#_{\mathcal{B}}(3)=\#_{\mathcal{B}}(4)=36$, the variation in dataset sizes across different configurations is solely determined by $\#_{\mathcal{C}}(n)$. Statistical results indicate that the training sizes are generally within the range of approximately 1.0M to 1.3M, without any substantial differences. Therefore, the training conditions across different experimental groups can be regarded as essentially fair.

\subsection{Model and Implementation}
\label{sec:model}

We adopted the vanilla architecture of LLaMA \citep{grattafiori2024llama}, a Transformer-based \citep{vaswani2017attention} decoder-only autoregressive language model.
Compared to the GPT-2 series \citep{radford2019language}, which more closely follows the vanilla Transformer architecture, the LLaMA architecture adopts RoPE \citep{su2021roformer} for positional encoding. This significantly enhances its ability to extrapolate to longer sequences, thereby helping to reduce the impact of potential differences in scene or query length during generalization evaluation.
We configured the model with 16 transformer layers, 24 self-attention heads, and an embedding dimension of 1024, resulting in approximately 373 million ($\sim$0.4B) parameters.

For each training session, model parameters were initialized using the Kaiming initialization method \citep{he2015delving}. Each training was conducted on a single 80G A100 GPU for 20 epochs. We employed a batch size of 512 and optimized the model using the AdamW optimizer \citep{loshchilov2017decoupled} with an initial learning rate set to 1e-4. To ensure stable training dynamics in the early stages, we introduced a linear warm-up over the first 1000 steps. Thereafter, a cosine annealing scheduler \citep{loshchilov2017sgdr} was applied to modulate the learning rate throughout the training process.




\bibliographystyle{unsrtnat}
\bibliography{sn-bibliography}

\newpage
\appendix
\part*{Supplementary Materials}

\newpage

\begin{algorithm}[htbp]
\caption{Graph Structure Generation}
\label{alg:1}
\KwIn{Number of nodes $n$, number of edges $m$}
\KwOut{Set $\mathcal{G}$ of pairs $(G,\pi)$ where $G$ is a valid DAG and $\pi$ is a topological order}
$\mathcal{G} \gets \emptyset$ \;
$V \gets \{0,1,\dots,n-1\}$ \;
\ForEach{permutation $\pi$ of $V$}{ 
    \textcolor{red}{\textbf{// Treat $\pi$ as a candidate topological order}}\;
    $E_{\text{cand}} \gets \emptyset$ \;
    \For{$i \gets 1$ \KwTo $n$}{
        \For{$j \gets i+1$ \KwTo $n$}{
            $E_{\text{cand}} \gets E_{\text{cand}} \cup \{(\pi_j,\pi_i)\}$ \; 
            \textcolor{red}{\textbf{// Only add backward edges w.r.t.\ $\pi$ to ensure acyclicity}}
        }
    }
    \ForEach{subset $E \subseteq E_{\text{cand}}$ with $|E|=m$}{
        \textcolor{red}{\textbf{// Enforce ``no isolated nodes''}} \;
        \If{all nodes in $V$ appear in some edge of $E$}{  
            Construct graph $G = (V,E)$ \;
            $\mathcal{G} \gets \mathcal{G} \cup \{(G,\pi)\}$ \;
        }
    }
}
\Return $\mathcal{G}$ \;
\end{algorithm}

\begin{algorithm}[htbp]
\caption{Character Order Generation over Graph Structures}
\label{alg:2}
\KwIn{Set $\mathcal{G}$ of pairs $(G,\pi)$ from \textbf{Algorithm \ref{alg:1}}}
\KwOut{Set $\mathcal{R}$ of pairs $(G,\texttt{character\_order})$}
\BlankLine
$\mathcal{R} \gets \emptyset$\;
\textcolor{gray}{\textbf{// process each graph $G$ with its topological order $\pi$}} \;
\ForEach{$(G,\pi) \in \mathcal{G}$}{ 
    \BlankLine
    \textcolor{gray}{\textbf{// build \texttt{parents} and \texttt{edge\_list} for current $G$}}\;
    \ForEach{$v \in V(G)$}{
        \texttt{parents[$v$]} $\gets$ empty list\;
    }
    \texttt{edge\_list} $\gets$ empty list\;
    \ForEach{$(u,v) \in E(G)$}{
        append $(u,v)$ to \texttt{edge\_list}\;
        append $u$ to \texttt{parents[$v$]}\;
    }

    \BlankLine
    \textcolor{red}{\textbf{// The topological order $\pi$ corresponds to the order in which characters exit the scene. During each traversal, the current node first enters the scene, and all of its parent nodes must also be ensured to enter simultaneously}}\;
    \texttt{character\_order} $\gets$ empty list\;
    \texttt{scene\_node} $\gets$ empty list\;
    \texttt{error} $\gets 0$\;

    \ForEach{$\texttt{node} \in \pi$}{ 
        \texttt{in\_list} $\gets$ \{ \texttt{node} \} if \texttt{node} $\notin \texttt{scene\_node}$ else $\emptyset$\;
        \ForEach{$n \in \texttt{parents[node]}$}{
            \If{$n \notin \texttt{scene\_node}$}{
                append $n$ to \texttt{in\_list}\;
            }
        }
        \ForEach{$n \in \texttt{in\_list}$}{
            \If{$n \notin \texttt{scene\_node}$}{
                append $n$ to \texttt{scene\_node}\;
            }
        }
        \texttt{out} $\gets$ \texttt{node}\;
        \textcolor{red}{\textbf{// Double-Check Graph Structure Validity: There must exist at least one edge between the characters currently present in the scene; otherwise, when a character leaves the scene, no valid interaction can be formed among them}}\;
        \ForEach{$n \in \texttt{scene\_node}$}{
            \If{$n \neq \texttt{out}$ \textbf{and} $(n,\texttt{node}) \notin \texttt{edge\_list}$}{
                \texttt{error} $\gets 1$\;
            }
        }
        remove \texttt{out} from \texttt{scene\_node}\;
        append \{\texttt{in}: \texttt{in\_list}, \texttt{out}: \texttt{out}\} to \texttt{character\_order}\;
    }

    \BlankLine
    \If{\texttt{error} = 0}{
        $\mathcal{R} \gets \mathcal{R} \cup \{(G,\texttt{character\_order})\}$\;
    }
}
\Return $\mathcal{R}$\;
\end{algorithm}

\begin{algorithm}[htbp]
\caption{Scene Template Generation over Character Orders}
\label{alg:3}
\KwIn{\\
Set $\mathcal{R}$ of pairs $(G,\texttt{character\_order})$ from \textbf{Algorithm \ref{alg:2}};\\
The number of containers $q$\\
}
\KwOut{\\
Set $\mathcal{S}$ of pairs $(G,\texttt{scene\_template})$\\
}

\BlankLine
\BlankLine
$\mathcal{S} \gets \emptyset$\;
\ForEach{$(G,\texttt{character\_order}) \in \mathcal{R}$}{
    $n \gets |V(G)|$\;
    \texttt{c\_label}$(v) \gets$ ``[$c_{v+1}$]''\;

    \texttt{text} $\gets$ ``There were ''\;
    \For{$j \gets 1$ \KwTo $q$}{
        \If{$j<q$}{\texttt{text} $\gets$ \texttt{text} + ``one [$b_j$], ''}
        \Else{\texttt{text} $\gets$ \texttt{text} + ``and one [$b_q$] in the room.\textbackslash n''}
    }
    \texttt{text} $\gets$ \texttt{text} + ``The [$a$] was initially in the [$\Tilde{b}$]\textbackslash n'' \;
    
    \ForEach{\texttt{step} $\in$ \texttt{character\_order}}{
        \texttt{in\_list} $\gets$ \texttt{step.in}\;
        \If{\texttt{in\_list} $\neq \emptyset$}{
            \texttt{names} $\gets$ join\big(\texttt{c\_label}$(v)$ for $v \in \texttt{in\_list}$, delimiter ``,''\big)\;
            \texttt{text} $\gets$ \texttt{text} + \texttt{names} $+\ $`` entered the room.\textbackslash n''\;
        }
        \texttt{out} $\gets$ \texttt{c\_label}(\texttt{step.out})\;
        \texttt{text} $\gets$ \texttt{text} + `` '' + \texttt{out} + `` moved the [$a$] to [$\Tilde{b}$].\textbackslash n'' \;
        \texttt{text} $\gets$ \texttt{text} + `` '' + \texttt{out} + `` exited the room.\textbackslash n''\;
    }

    \texttt{scene\_template} $\gets$ \texttt{text} \;
    $\mathcal{S} \gets \mathcal{S} \cup \{(G,\pi,\texttt{character\_order},\texttt{scene\_template})\}$\;
}
\Return $\mathcal{S}$\;
\end{algorithm}

\begin{figure}[htbp]
  \centering
  \includegraphics[width=1\columnwidth]{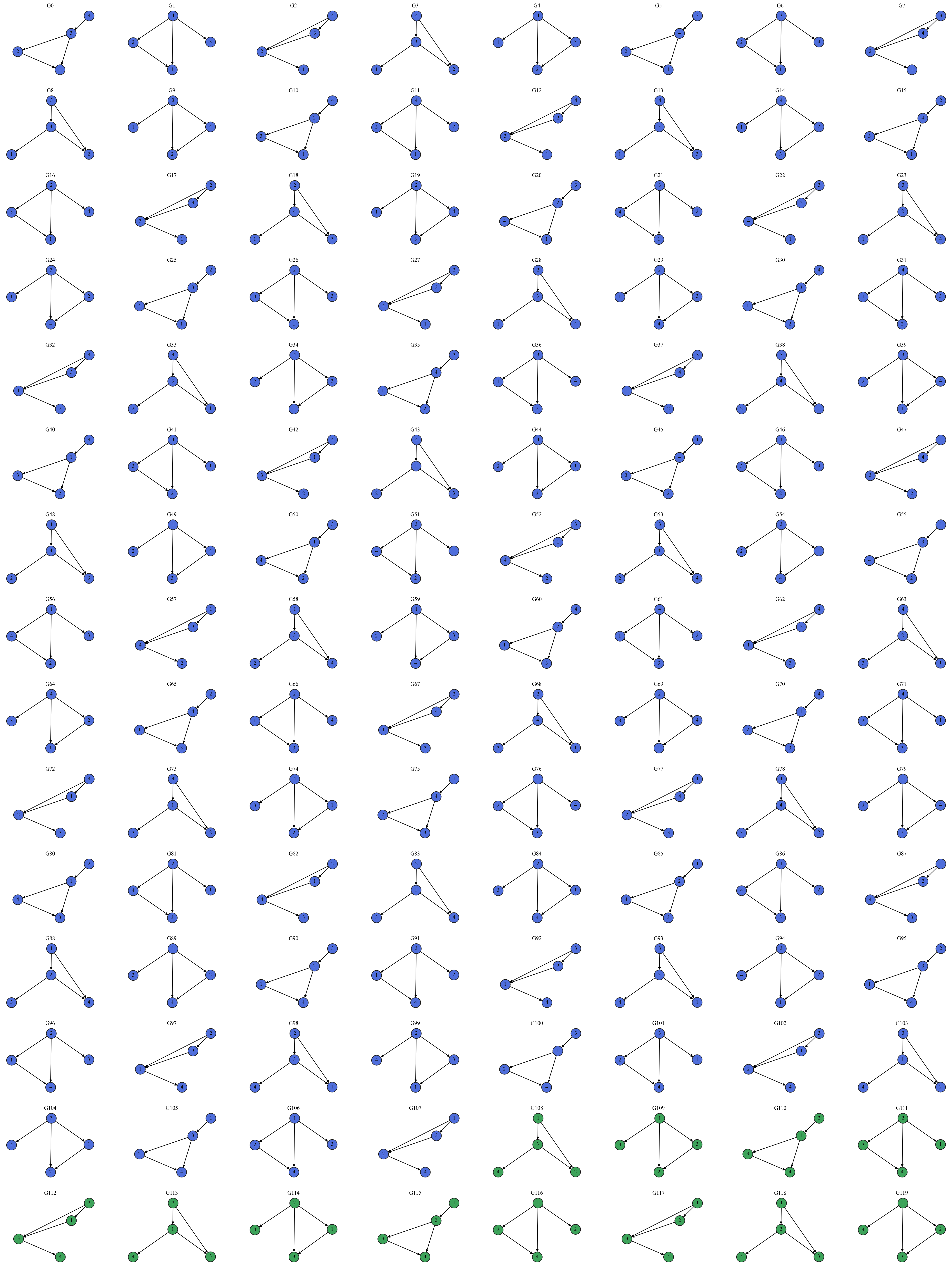}
  \caption{All graph structures used in this paper's experiments with {\bf 4 nodes and 4 edges}, where the blue graphs appear in the training data and the green graphs appear only in the test data.}
  \label{fig:graph45}
\end{figure}

\begin{figure}[htbp]
  \centering
  \includegraphics[width=1\columnwidth]{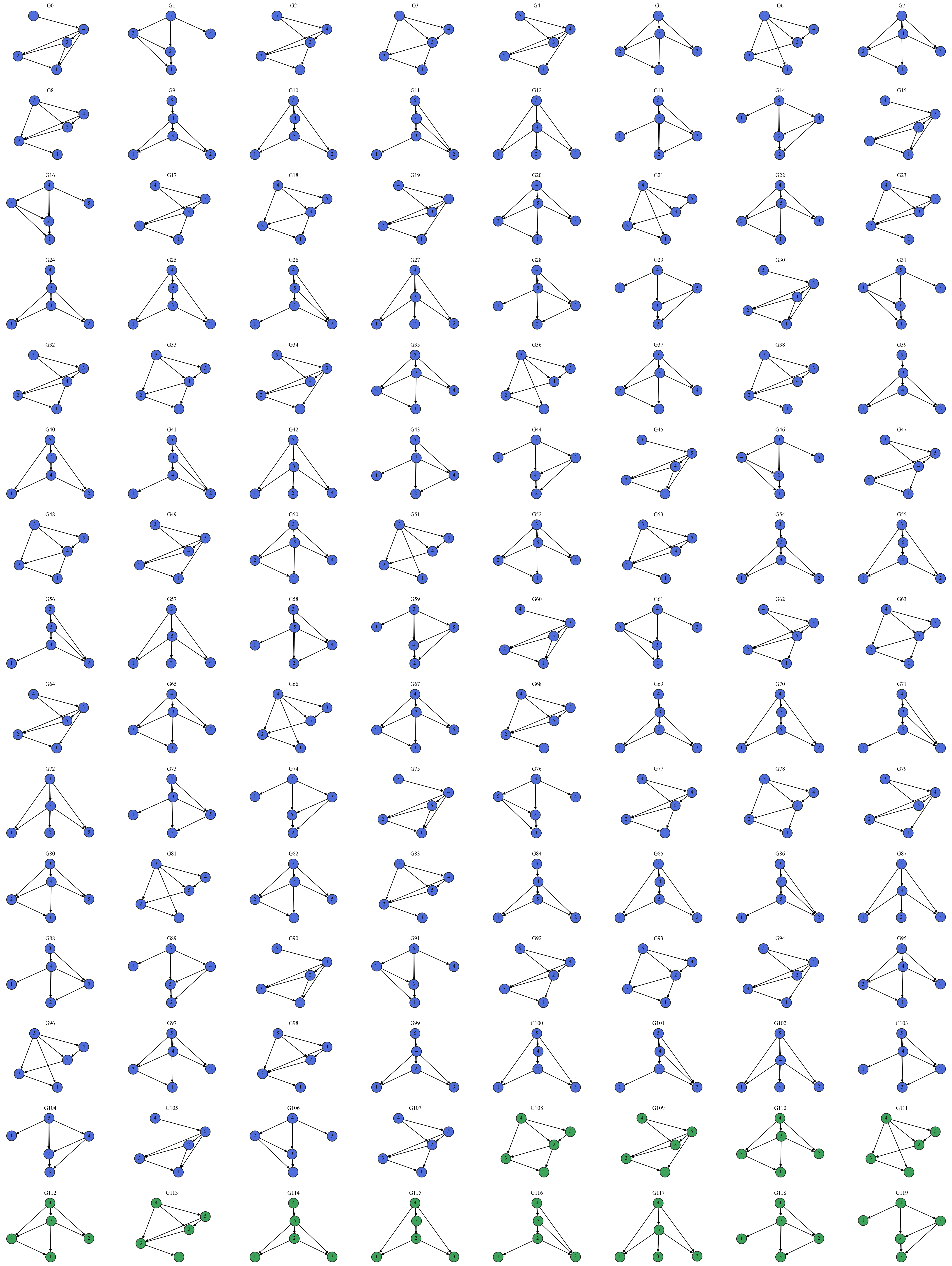}
  \caption{All graph structures used in this paper's experiments with {\bf 5 nodes and 7 edges}, where the blue graphs appear in the training data and the green graphs appear only in the test data.}
  \label{fig:graph57}
\end{figure}

\begin{figure}[htbp]
  \centering
  \includegraphics[width=1\columnwidth]{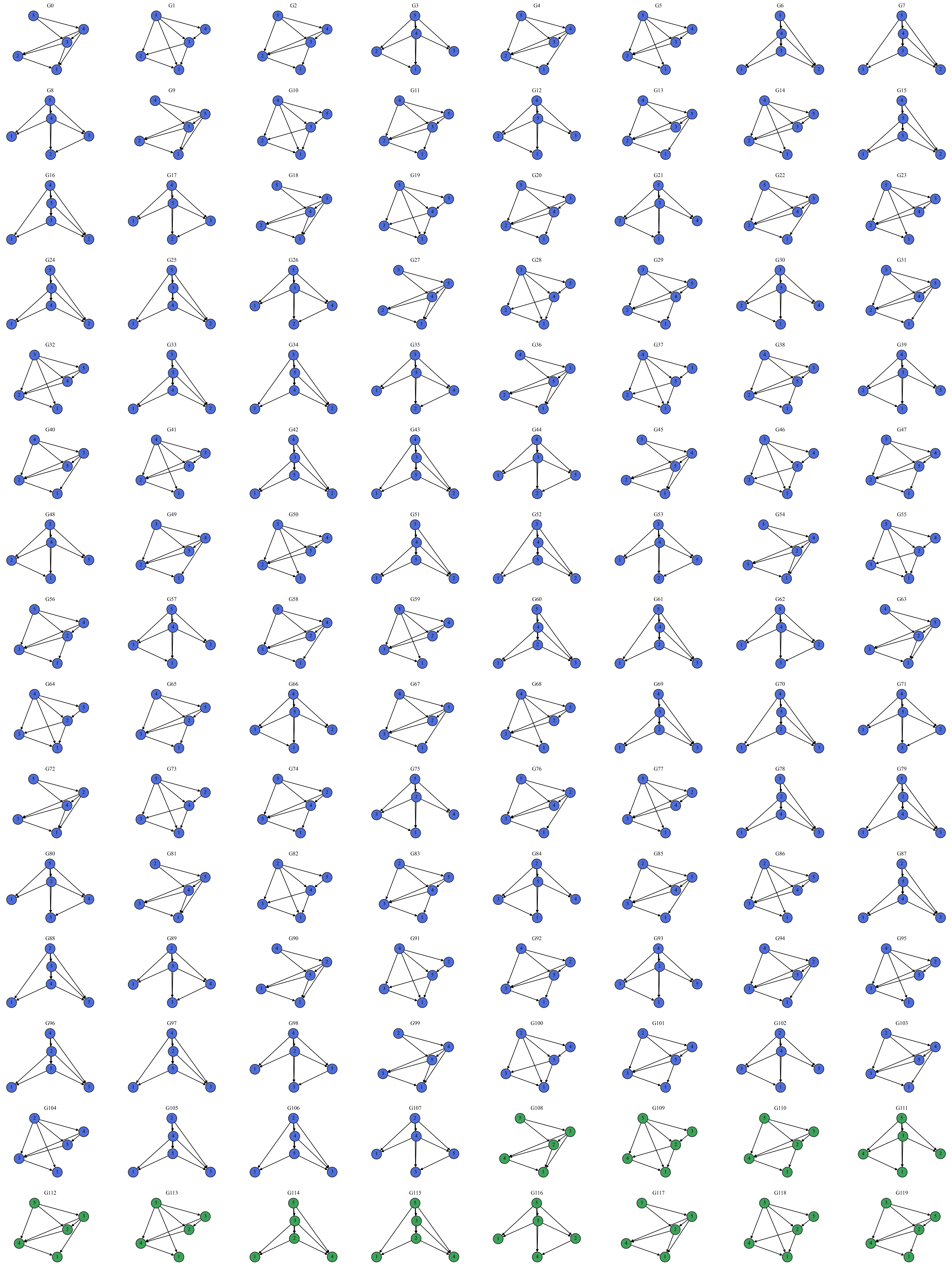}
  \caption{All graph structures used in this paper's experiments with {\bf 5 nodes and 8 edges}, where the blue graphs appear in the training data and the green graphs appear only in the test data.}
  \label{fig:graph58}
\end{figure}

\begin{figure}[htbp]
  \centering
  \includegraphics[width=1\columnwidth]{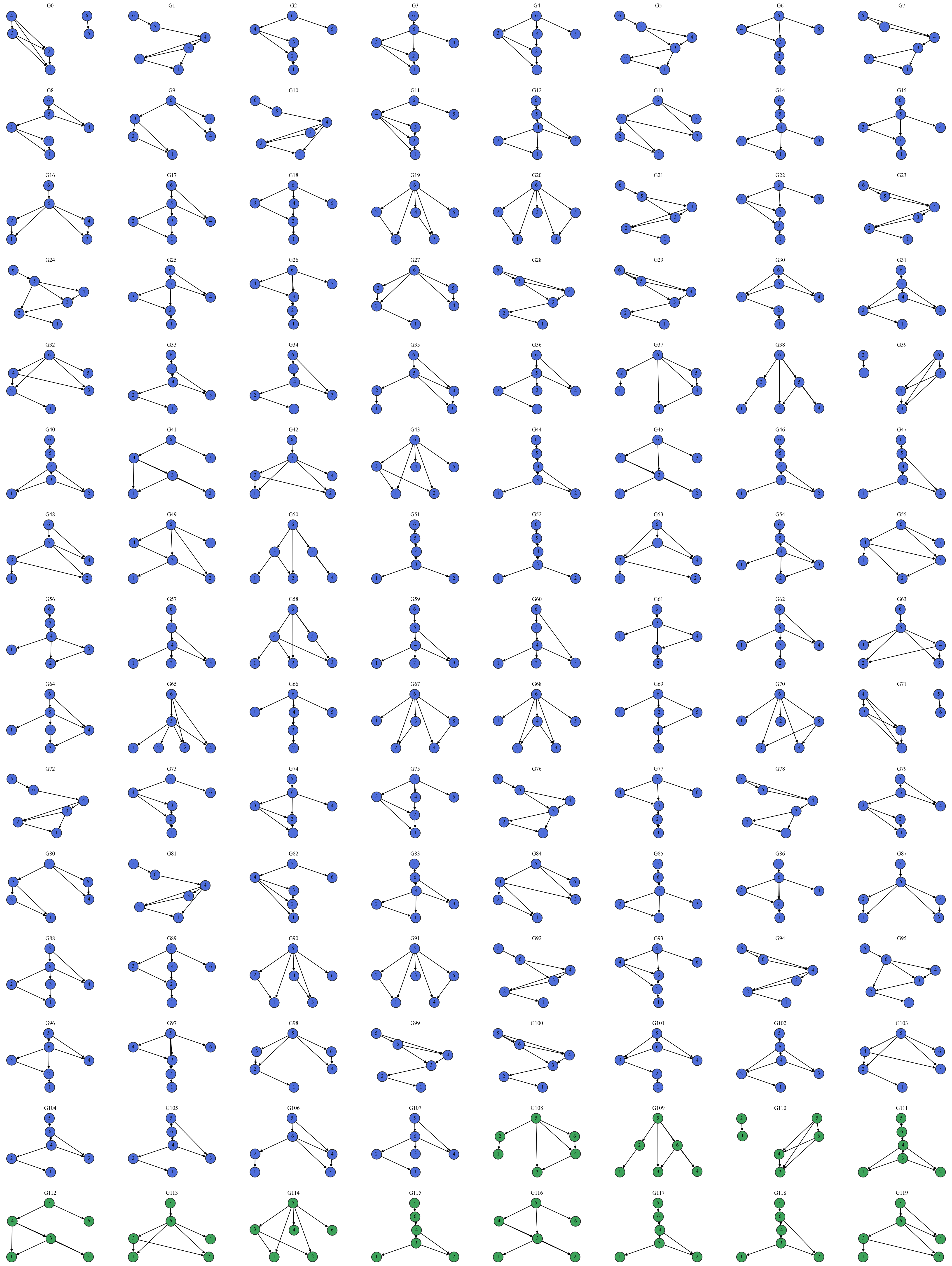}
  \caption{All graph structures used in this paper's experiments with {\bf 6 nodes and 7 edges}, where the blue graphs appear in the training data and the green graphs appear only in the test data.}
  \label{fig:graph67}
\end{figure}

\begin{figure}[htbp]
  \centering
  \includegraphics[width=1\columnwidth]{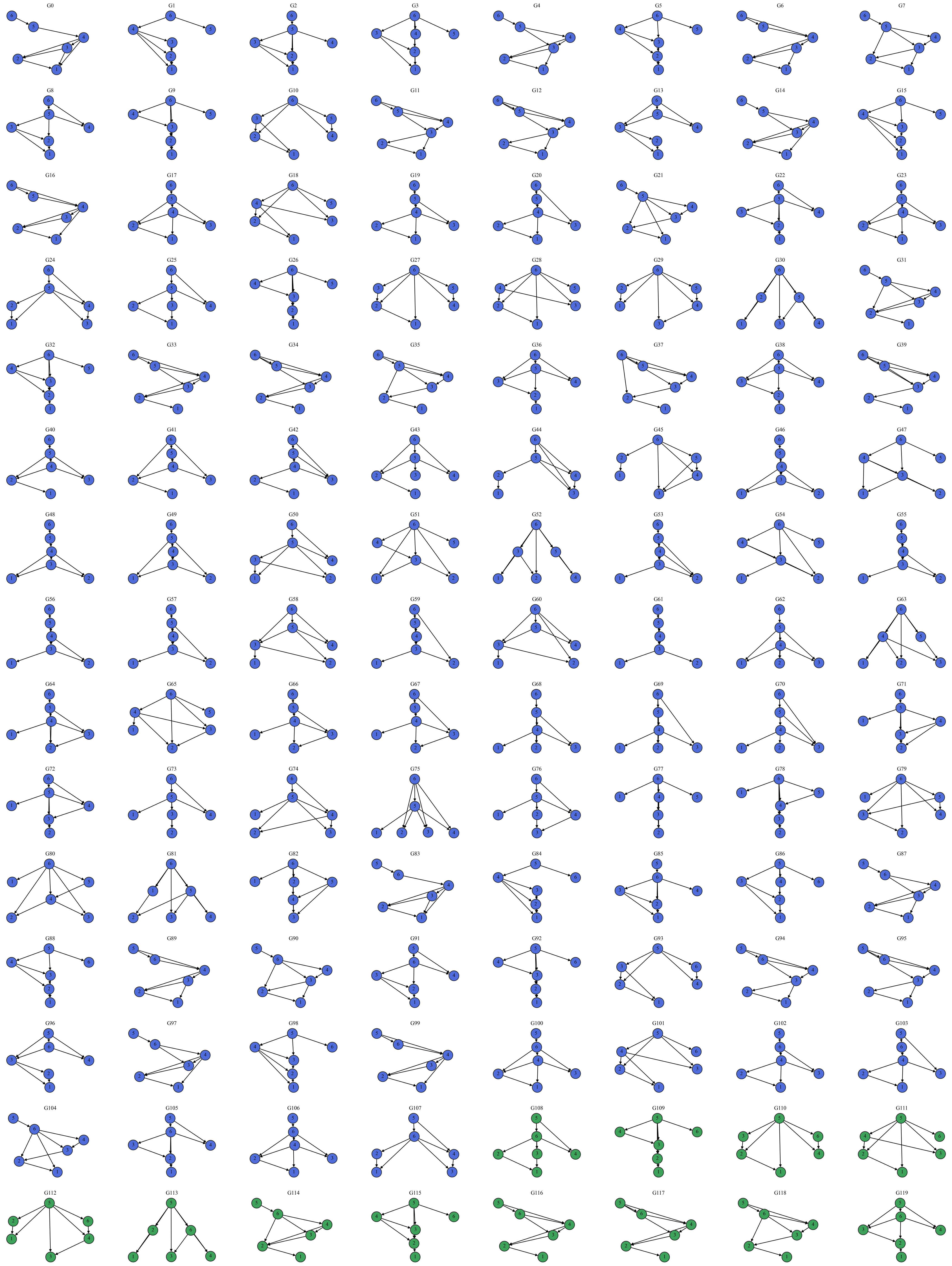}
  \caption{All graph structures used in this paper's experiments with {\bf 6 nodes and 8 edges}, where the blue graphs appear in the training data and the green graphs appear only in the test data.}
  \label{fig:graph68}
\end{figure}

\begin{figure}[htbp]
  \centering
  \includegraphics[width=1\columnwidth]{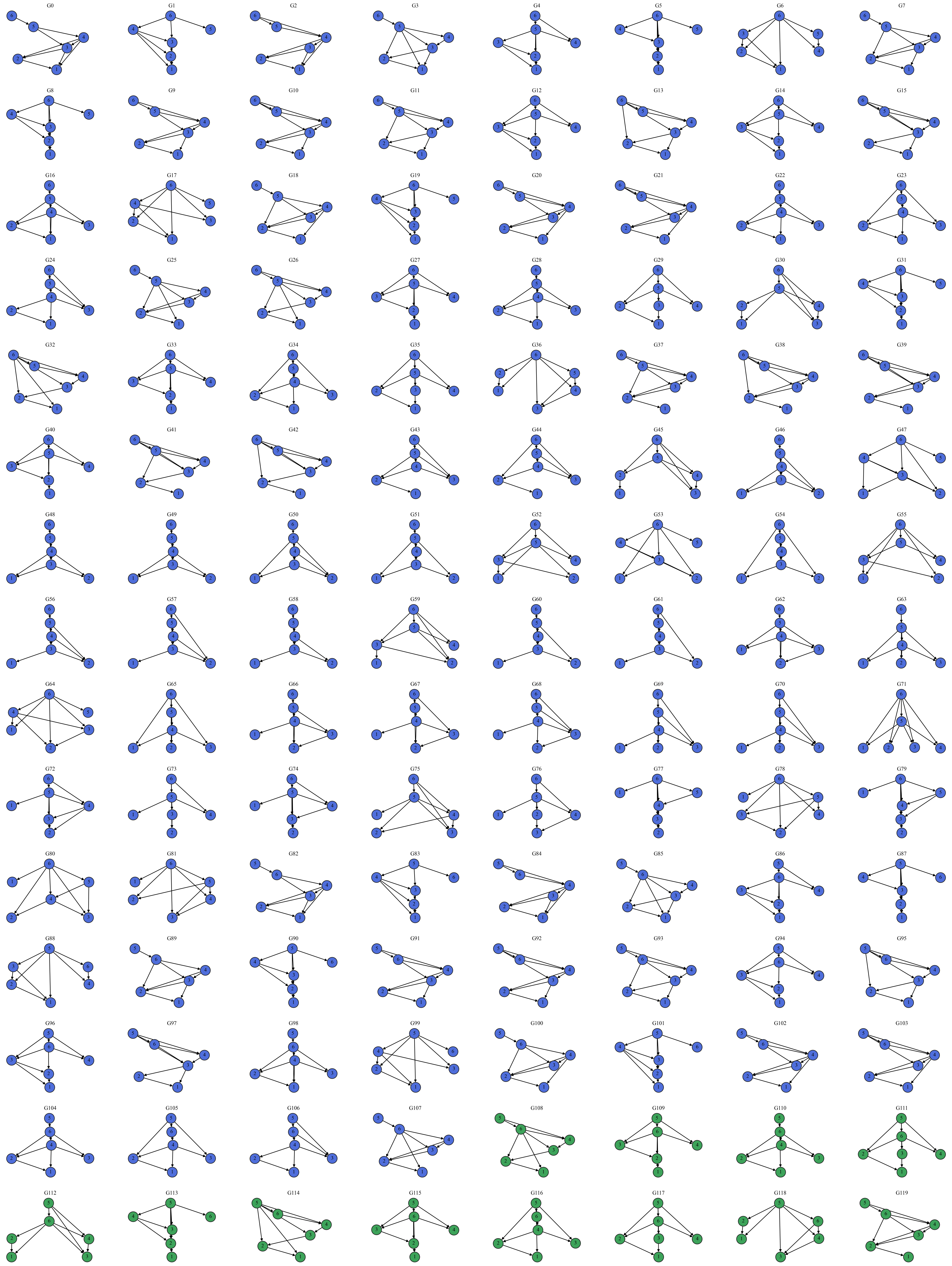}
  \caption{All graph structures used in this paper's experiments with {\bf 6 nodes and 9 edges}, where the blue graphs appear in the training data and the green graphs appear only in the test data.}
  \label{fig:graph69}
\end{figure}

\begin{figure}[htbp]
  \centering
  \includegraphics[width=1\columnwidth]{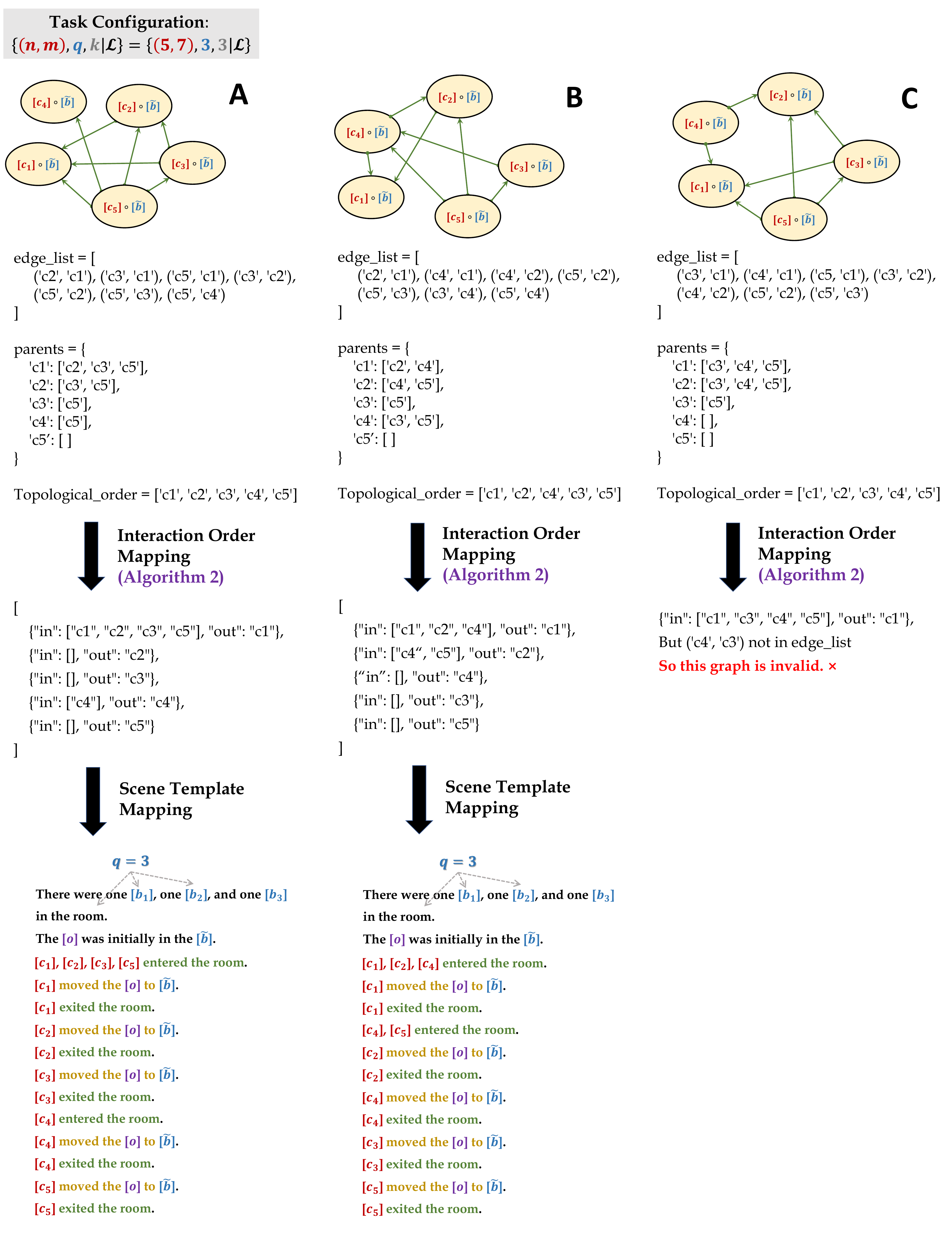}
  \caption{Illustrative process during the {\it scene template generation} stage under the configuration $\{(n,m), q, k\} = \{(5,7), 3, 3\}$, including verification and filtering of invalid graph structures.}
  \label{fig:mapping-case57}
\end{figure}

\begin{figure}[htbp]
  \centering
  \includegraphics[width=1\columnwidth]{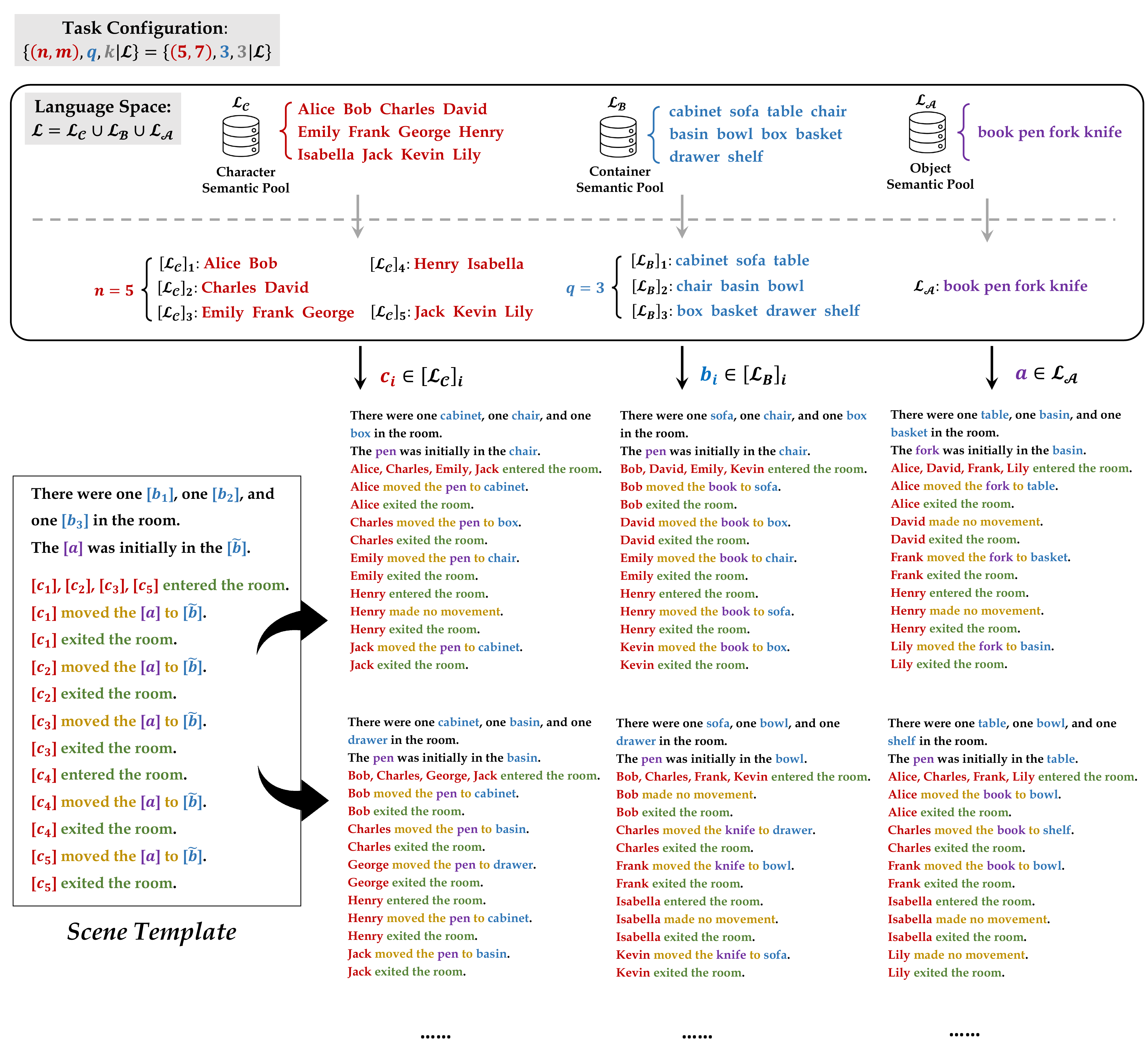}
  \caption{Illustrative process of semantic diversification during the {\it scene generation} stage under the configuration $\{(n,m), q, k\} = \{(5,7), 3, 3\}$.}
  \label{fig:semantic-case57}
\end{figure}

\begin{figure}[htbp]
  \centering
  \includegraphics[width=1\columnwidth]{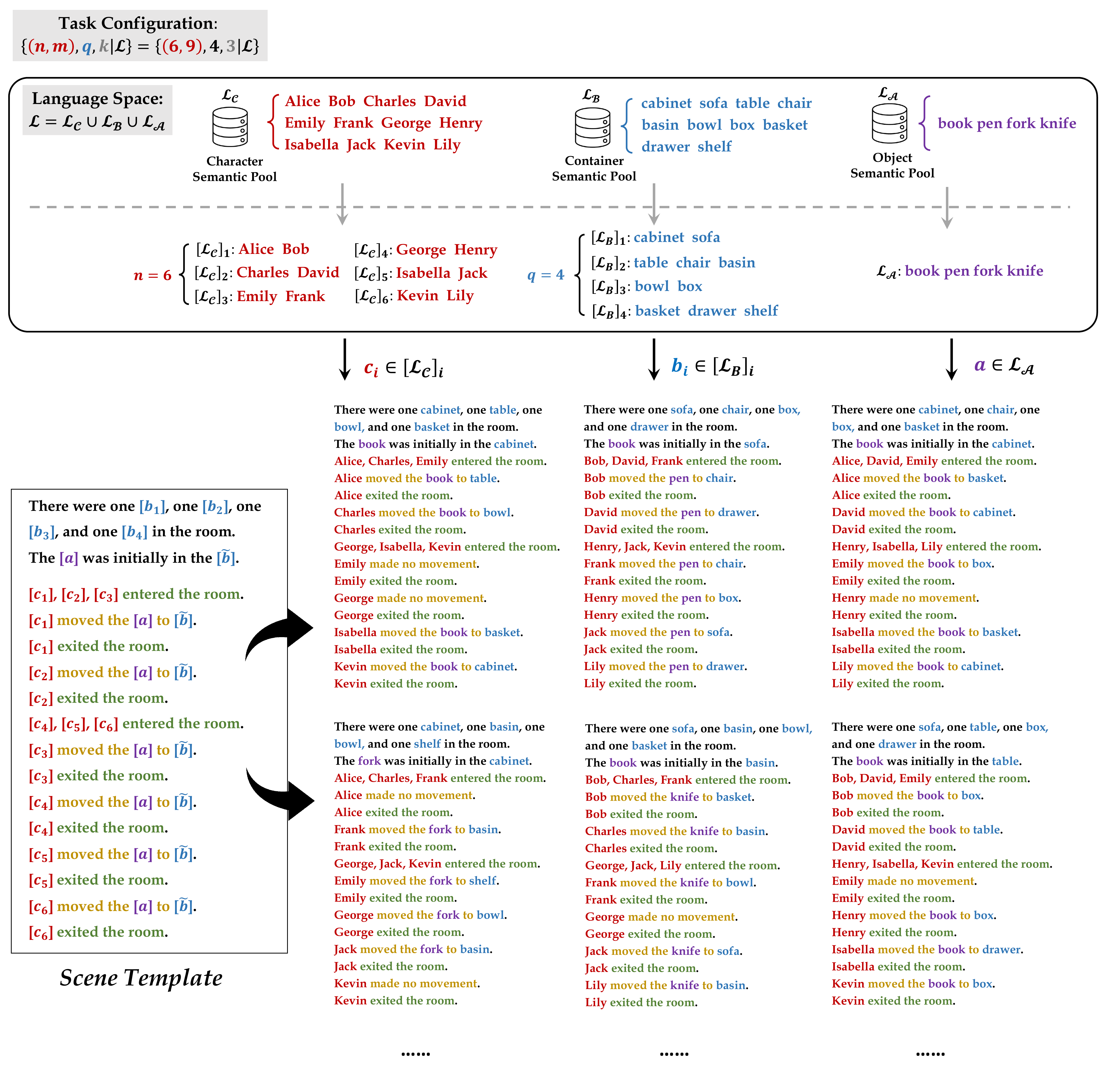}
  \caption{Illustrative process of semantic diversification during the {\it scene generation} stage under the configuration $\{(n,m), q, k\} = \{(6,9), 4, 3\}$.}
  \label{fig:semantic-case69}
\end{figure}













\end{document}